\documentclass[letterpaper]{article} 
\usepackage[submission]{aaai23}  
\usepackage{times}  
\usepackage{helvet}  
\usepackage{courier}  
\usepackage[hyphens]{url}  
\usepackage{graphicx} 
\usepackage{natbib}  
\usepackage{caption} 
\usepackage{algorithm}
\usepackage{algorithmic}

\usepackage{bbm}
\usepackage{amsfonts}
\usepackage{amsmath}
\usepackage{hyperref}
\hypersetup{breaklinks=true, colorlinks=true}
\usepackage{multirow, makecell}
\usepackage{subcaption}
\usepackage{tabularx}
\usepackage{csquotes}

\usepackage{newfloat}
\usepackage{listings}
\DeclareCaptionStyle{ruled}{labelfont=normalfont,labelsep=colon,strut=off} 
\floatstyle{ruled}
\newfloat{listing}{tb}{lst}{}
\floatname{listing}{Listing}
\title{CRAFT: Camera-Radar 3D Object Detection\\with Spatio-Contextual Fusion Transformer}
\author {
    Youngseok Kim\textsuperscript{\rm 1} \hspace{0.2cm}
    Sanmin Kim\textsuperscript{\rm 1} \hspace{0.2cm}
    Jun Won Choi\textsuperscript{\rm 2} \hspace{0.2cm}
    Dongsuk Kum\textsuperscript{\rm 1}
}
\affiliations {
    \textsuperscript{\rm 1}KAIST \hspace{0.6cm}
    \textsuperscript{\rm 2}Hanyang University\\
    \{youngseok.kim, sanmin.kim, dskum\}@kaist.ac.kr, junwchoi@hanyang.ac.kr
}

\begin{document}

\maketitle

\begin{abstract}
Camera and radar sensors have significant advantages in cost, reliability, and maintenance compared to LiDAR.
Existing fusion methods often fuse the outputs of single modalities at the result-level, called the late fusion strategy.
This can benefit from using off-the-shelf single sensor detection algorithms, but late fusion cannot fully exploit the complementary properties of sensors, thus having limited performance despite the huge potential of camera-radar fusion.
Here we propose a novel proposal-level early fusion approach that effectively exploits both spatial and contextual properties of camera and radar for 3D object detection.
Our fusion framework first associates image proposal with radar points in the polar coordinate system to efficiently handle the discrepancy between the coordinate system and spatial properties.
Using this as a first stage, following consecutive cross-attention based feature fusion layers adaptively exchange spatio-contextual information between camera and radar, leading to a robust and attentive fusion.
Our camera-radar fusion approach achieves the state-of-the-art 41.1\% mAP and 52.3\% NDS on the nuScenes test set, which is 8.7 and 10.8 points higher than the camera-only baseline, as well as yielding competitive performance on the LiDAR method.
\end{abstract}

\section{Introduction}
3D object detection is an essential task for autonomous driving as well as mobile robots.
Thanks to the low-cost, high-reliability, and low-maintenance of camera and radar sensors, they are already deployed to a large number of mass production vehicles for active safety systems.
Moreover, the rich semantic information of the camera and long-range detection with weather condition robustness of radar are essential attributes that LiDAR cannot provide.
Nevertheless, learning-based camera-radar 3D object detection~\cite{lim2019radar, kim2020grif, nabati2021centerfusion} for autonomous driving has not been thoroughly explored.

\setcounter{figure}{0}
\begin{figure}[t]
\begin{center}
\includegraphics[width=0.92\columnwidth]{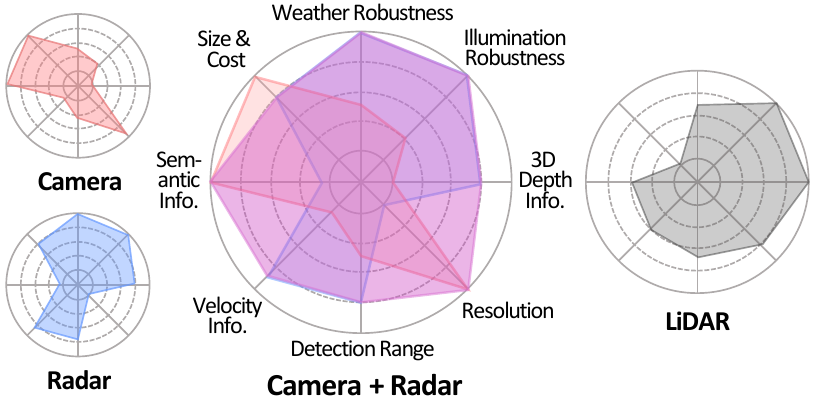}
\end{center}
\caption{
Sensor characteristics of camera, radar, and LiDAR.
Camera-radar fusion has high potential considering spatially and contextually complementary properties.
}
\label{fig:intro}
\end{figure}

Existing camera-radar fusion methods often use a late fusion strategy that fuses detection-level outputs of separated camera and radar detection algorithms with heuristic logic.
Such methods can have the benefit of exploiting the off-the-shelf detection algorithms that are independently developed by automotive suppliers as modular components.
However, late fusion strategies that rely on heuristics and post-processing techniques suffer from performance-reliability trade-offs, especially when two sensor predictions disagree; thus, late fusion cannot exploit the full potential of each sensor.
In contrast, a learning-based early fusion strategy that fuses the intermediate information of each sensor at an early stage has much higher potential, but it requires an in-depth understanding of each sensor's characteristics to find the ``optimal'' way to fuse.

However, it is non-trivial to develop early fusion for camera and radar due to the unique characteristics of each sensor, as illustrated in Fig. \ref{fig:intro}.
Camera-LiDAR early fusion methods, which are relatively well-studied, fuse point-wise features~\cite{vora2020pointpainting, huang2020epnet} by associating image pixel with projected LiDAR point or fuse image and LiDAR proposal features~\cite{ku2018joint} by extracting a region of interests (RoIs) with RoIPool~\cite{Sun2015}.
However, adapting camera-LiDAR fusion methods for camera-radar is unsuitable since those strategies are established by accurate LiDAR measurement ($\pm2cm$), but radar has low accuracy and measurement ambiguities.
Due to the nature of the radar mechanism, the radar has high resolution and accuracy in the radial direction ($0.4m$ and $\pm0.1m$), which is measured by Fast Fourier Transformation FFT.
Meanwhile, the azimuthal measurement obtained by digital beamforming using multiple receive antennas~\cite{johnson1992array} is inaccurate ($4.5^{\circ}$ and $\pm1^{\circ}$, approximately $4m$ and $\pm1m$ at 50$m$ distance).
Measurement ambiguities are the cases when radar points are occasionally missed on objects (false negatives due to the low radar cross-section) or shown in the background (false positives due to radar multi-path problem), thus these have to be considered.
The camera, meanwhile, has very complementary spatial characteristics to radar~\cite{ma2021delving, hung2022let}.
Dense camera pixels provide accurate azimuthal resolution and accuracy, but the depth information is not provided due to the perspective projection.

Motivated by the aforementioned challenges of camera and radar, camera-radar fusion has to be able to 1) robustly operate even radar points are not reflected from the object and 2) effectively complement spatial (range/azimuth) and contextual (semantic/Doppler) information.
To this end, we present Camera RAdar Fusion Transformer (CRAFT), a robust and effective fusion framework for 3D object detection.
We propose the soft association strategy between image proposal and radar points and cross-attention based feature fusion method, which can effectively exploit spatial characteristics and robustly handle radar measurement ambiguity.
Specifically, we first extract camera and radar features by modality-specific feature extractors (\textit{i.e.}, DLA-34~\cite{Yu2018}, PointNet++~\cite{qi2017pointnet}) and predict 3D objects with a lightweight camera 3D object detector.
Given image proposals and radar points, we associate image proposal with radar points by Soft Polar Association, which queries points within the uncertainty-aware adaptive ellipsoidal threshold in the polar coordinate.
To deal with the wrongly associated radar points (background clutters), we then attentively fuse the image proposal feature and radar point features by consecutive cross-attention based encoder layers.
Our Spatio-Contextual Fusion Transformer can effectively exchange spatial and contextual information and adaptively determine where and what information to use for fusion.
Finally, the class-specific decoder head predicts fusion score and offsets in the polar coordinate to refine the image proposal.
Since some objects do not have valid radar points, fused output with a low fusion score is discarded, and the image proposal is used as the final prediction.

To summarize, our contributions are as follows:

\begin{itemize}
    \item We investigate the characteristics of camera-radar and propose a proposal-level early fusion framework that can mitigate the coordinate system discrepancy and measurement ambiguities.
    \item We propose Soft Polar Association and Spatio-Contextual Fusion Transformer that effectively exchanges complementary information between camera and radar features.
    \item We achieve the state-of-the-art 41.1\% mAP and 52.3\% NDS on nuScenes test set, which significantly boosts the camera-only baseline with a marginal additional computation cost.
    \end{itemize}

\section{Related Work}

\subsection{Camera and Radar for 3D Object Detection}
With an advance in 2D object detection~\cite{Sun2015, tian2019fcos}, image view approaches extend a 2D detector with additional 3D regression branches.
CenterNet~\cite{Zhou2019} and FCOS3D~\cite{wang2021fcos3d} directly regress depth to the object and 3D size from its center without using an anchor, which is later improved in PGD~\cite{wang2021probabilistic} by utilizing geometric prior.
Image view approaches can well exploit GPU-friendly operations and therefore runs fast but suffer from inaccurate depth estimation, which is the naturally ill-posed problem.

Another approaches are to transform image features in the perspective view into a top-down view and predict 3D bounding boxes from the BEV features using LiDAR detection heads~\cite{lang2019pointpillars, yin2021center}.
CaDDN~\cite{reading2021categorical} and LSS~\cite{philion2020lift} generate the BEV features using the depth distribution, and PETR~\cite{liu2022petr} and BEVFormer~\cite{li2022bevformer} generate the BEV features by using pre-defined grid-shaped BEV queries to image features.
Such BEV methods can utilize more depth information during training, thus showing better localization performance, but operates slow.

Radar data can have various representations such as 2-D FFT~\cite{lin2018human}, Range-Azimuth-Doppler Tensor~\cite{major2019vehicle, kim2020low}; and a radar point cloud~\cite{Caesar2020, meyer2019automotive} is an affordable representation for autonomous driving applications.
Although the radar point cloud is similar to LiDAR, naively adopting LiDAR methods to radar is inappropriate.
\cite{ulrich2022improved} adapt PointPillars with KPConv~\cite{thomas2019kpconv} and Radar-PointGNN~\cite{svenningsson2021radar} adapt GNN to radar, but their performances are much inferior to LiDAR methods.
Although the high potential of radar, radars have not yet been thoroughly investigated in autonomous driving.

\subsection{Camera-LiDAR for 3D Object Detection}
Camera-LiDAR fusion has gained significant interest for 3D detection, and existing approaches can be roughly classified into result-, proposal- and point-level fusion.
Thanks to the rich intermediate feature representation of the learning-based method, a line of work~\cite{ku2018joint, yoo20203d} fuses RoI proposal features to complement two modalities' information.
Recent works~\cite{vora2020pointpainting, bai2022transfusion} further propose point-level fusion that can fuse more fine-grained features at an earlier stage.
However, methods using LiDAR are established on the strong assumption that LiDAR measurement and camera-LiDAR calibration is accurate, which is not valid under a radar setting.

\setcounter{figure}{1}
\begin{figure*}[t]
\begin{center}
\includegraphics[width=0.86\textwidth]{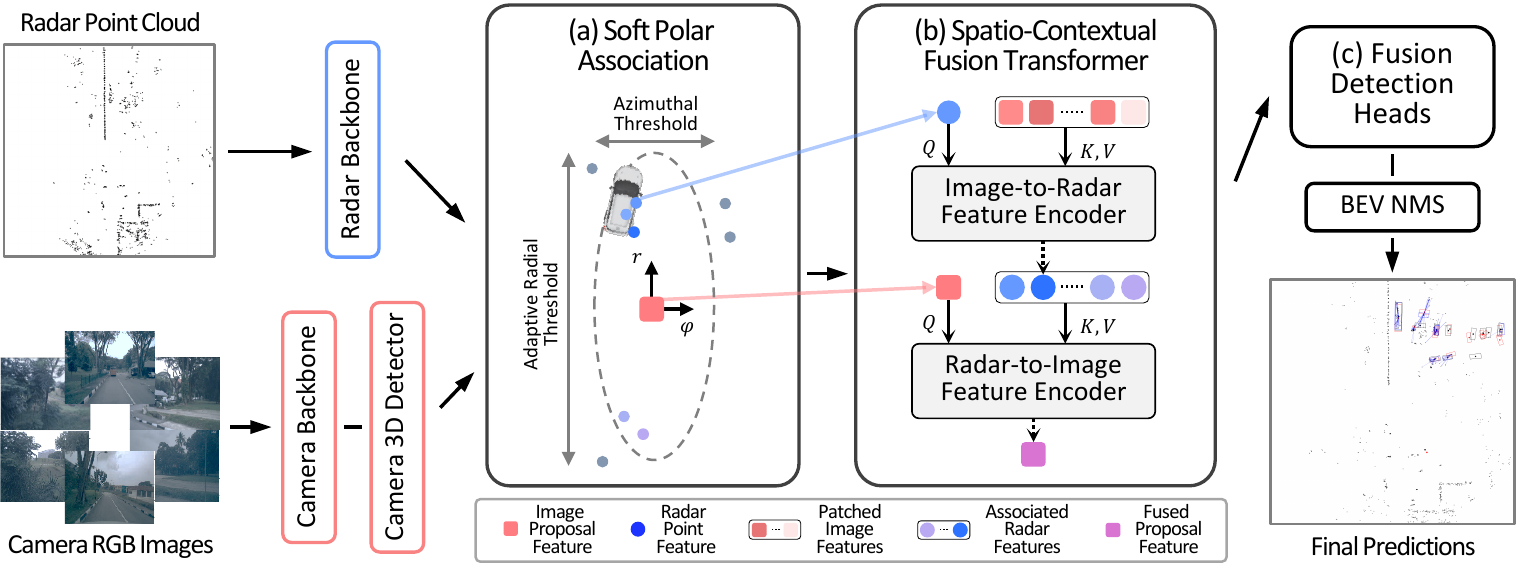}
\end{center}
\caption{
Overall architecture of CRAFT.
Modality-specific backbones extract features, and camera 3D detector predicts 3D image proposals.
(a) Given the image proposals, we associate radar points around the image proposals using adaptive thresholds in polar coordinates.
(b) Then, consecutive cross-attention layers adaptively complement spatial and contextual information of camera and radar features.
(c) Finally, fusion detection heads predict fusion score and offset to refine the image proposal and to output final predictions.
}
\label{fig:architecture}
\end{figure*}

\subsection{Camera-Radar Fusion}
Only a few methods attempt to detect 3D objects using camera radar fusion in autonomous driving.
\cite{lim2019radar} transforms the camera image into BEV using inverse perspective mapping, assuming a planar road scene such as a highway to mitigate the coordinate discrepancy between camera and radar.
\cite{meyer2019deep, kim2020low} adapt AVOD~\cite{ku2018joint} to fuse front view camera image with point- and tensor-like radar data, respectively.
GRIF Net~\cite{kim2020grif} further proposes a gating mechanism to adaptively fuse camera and radar RoI features handling the radar sparsity.
CenterFusion~\cite{nabati2021centerfusion} lifts image proposals into a 3D frustum and associates a single closest radar point inside RoI to fuse.
Thus, existing camera-radar 3D detectors have not thoroughly considered the spatial properties of camera and radar.

Some literature exploits camera and radar for depth completion tasks considering radar sparsity, inaccurate measurement, and ambiguity.
\cite{lin2020depth} proposes a radar noise filtering module using a space-increasing discretization (SID) threshold to filter outlier noises, then uses filtered radar points to improve depth prediction.
RC-PDA~\cite{long2021radar} further proposes pixel depth association to find the one-to-many mapping between radar point and image pixels that filter and densify radar depth map.
In our work, we explore a robust fusion method using an attention mechanism to handle these limitations of radar effectively.

\section{Method}
In the camera-radar fusion-based 3D detection task, we take surrounding images and radar point clouds with corresponding intrinsics and extrinsics as input.
Given camera feature maps from the backbone, camera detection heads first generate object proposals with a 3D bounding box.
A \textit{Soft Polar Association (SPA)} module then associates radar points with object proposals using adaptive thresholds in the polar coordinate to handle the spatial characteristics of sensors effectively.
Further, a \textit{Spatio-Contextual Fusion Transformer (SCFT)} fuses camera and radar features to complement spatial and contextual information of a single modality to another.
Finally, \textit{Fusion Detection Heads} decode the fused object features to refine initial object proposals.
An overall framework is provided in Fig. \ref{fig:architecture}.

\subsection{Backbones and Camera 3D Object Detector}
Given $N$ camera images $I\in {\mathbb{R}}^{N\times H\times W\times 3}$, the 3D detector aims to classify and localize objects with a set of 3D bounding boxes.
Specifically, we feed multi-camera images to the backbone network (\textit{e.g.}, DLA-34~\cite{Yu2018}) and obtain downsampled image features $F \in {\mathbb{R}}^{N\times h\times w\times C}$ of multiple camera views.
Taking camera view features $F$ as input, the convolutional detection heads predict the projected 3D center $(u,v)$ (called keypoint), depth $d$, and other attributes in the image plane.
We additionally predict the variance of depth $\sigma$ following~\cite{kendall2017} to represent the uncertainty of depth regression.
Keypoints are transformed into the 3D camera coordinate system with camera intrinsics $({x^{\prime}},\ {y^{\prime}}, f_u, f_v)$ as follows:

\begin{equation}
z_c=d, x_c=\frac{(u-{x^{\prime}})\times z}{f_u}, y_c=\frac{(v-{y^{\prime}})\times z}{f_v},
\label{eq:1}
\end{equation}
then 3D bounding boxes in each camera coordinate system are transformed into the vehicle coordinate system.
The bounding box $b=(x,y,z,\sigma,w,l,h,\theta,v)$ consists of the 3D center location $(x,y,z)$ with depth variance $\sigma$, 3D dimension $(w,l,h)$, yaw orientation $\theta$, and velocity $v$.
In the keypoint detection setting~\cite{Zhou2019}, the feature at the keypoint $f_{n,u,v} \in {\mathbb{R}}^{C}$ implicitly represents the object.
Thus, we denote a set of $M$ image proposals $\{i_m\}_{m=1}^M$, where $i_m =\left[b_m,f_m^i \right]$ indicates the bounding box and its features.
Note that proposing a camera 3D detector is out of the scope of this paper, and we expect that other research progress~\cite{wang2021fcos3d, park2021pseudo} could further improve our results.

For a set of $K$ radar points $\{p_k = \left[v_k,f_k \right] \}_{k=1}^K$ with 3D location $v_i \in \mathbb{R}^3$ and properties $f_i \in \mathbb{R}^F$ (\textit{e.g.}, radar cross-section (RCS), ego-motion compensated radial velocity), we extract the high-dimensional features $f_k^r \in {\mathbb{R}}^{C}$ using a point feature extractor (\textit{e.g.}, PointNet++~\cite{qi2017pointnet++}).
As the number of radar points $K$ is considerably small, we discard subsampling strategies such as farthest point sampling (FPS)~\cite{shi2019pointrcnn, yang20203dssd}.
The detailed architectures for the camera 3D detector and radar backbone are provided in Appendix \ref{chap:appendix A}.

\setcounter{figure}{2}
\begin{figure}[t]
\begin{center}
\includegraphics[width=0.9\columnwidth]{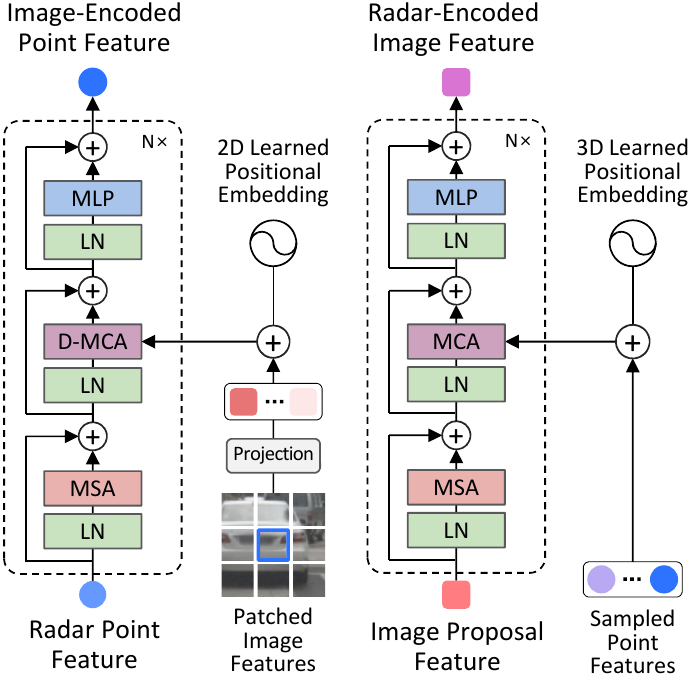}
\end{center}
\caption{
Architectures of the Image-to-Radar (left) and Radar-to-Image (right) Feature Encoder.
Radar and image features exchange spatial and contextual information by consecutive feature encoders.
We use RGB image for patched image features only for visualization.
}
\label{fig:FE}
\end{figure}

\subsection{Soft Polar Association}

Specifically, we calculate eight corners from the image proposal bounding box $b$ and transform corners and radar points in Cartesian coordinate to polar coordinate $O^{(P)} = \{(r,\phi,z)_{m,j}\}_{m,j=1}^{M,8}, v^{(P)} = \{(r,\phi,z)_k\}_{k=1}^K$.
Inspired by Ball Query~\cite{qi2017pointnet++}, we find a subset of radar points $\{p^{m}=[v_i, f_i]\}_{i=1}^{K^{'}}$ for each object proposal within a certain distance to the image proposal but with adaptive thresholds.

\begin{equation}
\phi_{m,l} < \phi_{i} < \phi_{m,r}
\label{eq:azimuthal thrheshold}
\end{equation}

\begin{equation}
r_{m,f} - (\gamma + \sigma \frac{r_{c}}{\delta}) < r_{i} <  r_{m,b} + (\gamma + \sigma \frac{r_{c}}{\delta})
\label{eq:radial threshold}
\end{equation}
Eq. \ref{eq:azimuthal thrheshold} is the azimuthal threshold for radar points in $p^m$ where $\phi_{m,l}$ and $\phi_{m,r}$ denote the angle of the most left and right corner points in polar coordinate.
The radial threshold is represented in Eq. \ref{eq:radial threshold}, where $r_{m,f}, r_{m,b}$ refer to radial distances of most front and back corner points of the image proposal, $\gamma$ is a minimum range, $\delta$ is a hyper-parameter to modulate the radius size, and $\sigma$ is the depth variance of the image proposal. 
$r_{c} = \frac{r_{m,f} + r_{m,b}}{2}$ denotes the radial distance of the center point of the image proposal.
In this way, the proposed association using an adaptive threshold can maximize the chance of using informative foreground radar points while excluding most clutter points in the azimuthal direction.


\subsection{Spatio-Contextual Fusion Transformer}
In our camera-radar fusion framework, the primary role of radar is to refine the location of the image proposal if there are radar returns from the object.
The Spatio-Contextual Fusion Transformer (SCFT) aims to exchange the spatial and contextual information between camera and radar, but determining which image pixels corresponding to the radar point is a difficult problem.
Thus we design fusion modules with cross-attention layers to make the fusion network learn where and what information should be taken from image and radar.

\subsubsection{Image-to-Radar Feature Encoder}
For spatio-contextual fusion, Image-to-Radar Feature Encoder (I2R FE) first provides the semantic information from image to radar.
Each radar point is projected to the image plane, and then a patch is defined around the projected location.
Instead of setting a fixed size patch, we design the size of the patch to be a function of radar distance so that the radar points can attend to a wider region when the object is closer, which takes more pixels in the image.
The adaptive size is similar to the inverse function of space-increasing discretization (SID)~\cite{Fu2018}:
\begin{equation}
\tau(d) = \lfloor W exp(-d/\beta + \alpha) \rfloor,
\label{eq:2}
\end{equation}
where we heuristically set $W=3.5$, $\alpha=2$, and $\beta=55$.
Then, $\tau \times \tau$ size patched image feature map is resized to the fixed size $w \times w$ using bilinear interpolation.

Inspired by Deformable DETR~\cite{zhu2020deformable}, we adopt a deformable multi-head cross-attention (D-MCA) module to attend to a small set of key sampling points (patched image features) around a reference point (projected radar pixel).
Given a radar query feature $f_k^r \in\mathbb{R}^{C}$, I2R FE adaptively extracts features from the patched image features $f_k^i \in\mathbb{R}^{w \times w \times C}$ even radar point is not accurately projected into the image.
However, the attention network desires proper supervision to learn which image feature is informative for the fusion effectively.
Accordingly, we design an auxiliary task for each image-encoded radar feature to predict the probability that the radar point is inside the 3D bounding box.

\subsubsection{Radar-to-Image Feature Encoder}
Following Radar-to-Image Feature Encoder (R2I FE) provides the spatial information of radar points to image proposal.
Unlike I2R FE operates in a 2D camera coordinate, R2I FE takes inputs in a polar coordinate since a regression target of 3D object detection is to predict the location in 3D space.
Following Transformer methods designed for 3D point cloud~\cite{pan20213d, misra2021end}, we use a cross-attention~\cite{vaswani2017attention} with 'pre-norm' techniques~\cite{xiong2020layer}.
Specifically, R2I FE takes the image proposal feature query $f_m^i \in \mathbb{R}^{C}$ and a set of radar point features $f_{k^{'}}^r \in \mathbb{R}^{K^{'}\times C}$ as input, then produces a radar-encoded image proposal feature that is later used to refine the proposal.
We additionally add a batch of zeros to the key-value sequences (\texttt{add$\_$zero$\_$attn} in PyTorch \texttt{nn.MultiHeadAttention}) so that attention can be assigned to it when none of the associated radar points are reflected from the object.

\setlength{\tabcolsep}{0.6em}
\setcounter{table}{0}
\begin{table*}[!t]
    \caption{
    State-of-the-art comparison on nuScenes \texttt{test} set. 
    `L', `C', and `R' represent LiDAR, camera, and radar, respectively.
    $*$ are initialized with FCOS3D~\cite{wang2021fcos3d} checkpoint and $\dagger$ use image flipping test time augmentation.
    }
    \begin{center}
    \resizebox{0.88\textwidth}{!}{
    \begin{tabular}{c|c|c||cc|ccccc|c}
        \hline
        Method & Input & Backbone & NDS$\uparrow$ & mAP$\uparrow$ & mATE$\downarrow$ & mASE$\downarrow$ & mAOE$\downarrow$ & mAVE$\downarrow$ & mAAE$\downarrow$ & FPS \\
        \hline
        PointPillars (\citeyear{lang2019pointpillars})    & L & - & 45.3 & 30.5 & 0.517 & 0.290 & 0.500 & 0.316 & 0.368 & 61 \\
        CenterPoint (\citeyear{yin2021center})           & L & - & 67.3 & 60.3 & 0.262 & 0.239 & 0.361 & 0.288 & 0.136 & 30 \\
        \hline
        Radar-PointGNN (\citeyear{svenningsson2021radar}) & R & - &   3.4 & 0.5 & 1.024 & 0.859 & 0.897 & 1.020 & 0.931 & - \\
        KPConvPillars (\citeyear{ulrich2022improved})     & R & - &  13.9 & 4.9 & 0.823 & 0.428 & 0.607 & 2.081 & 1.000 & - \\
        \hline
        CenterNet (\citeyear{Zhou2019})                     & C & HGLS & 40.0 & 33.8 & 0.658 & 0.255 & 0.629 & 1.629 & 0.142 & - \\
        FCOS3D$^\dagger$ (\citeyear{wang2021fcos3d})                  & C & R101 & 42.8 & 35.8 & 0.690 & 0.249 & 0.452 & 1.434 & 0.124 & 1.7 \\
        PGD$^\dagger$ (\citeyear{wang2021probabilistic})              & C & R101 & 44.8 & 38.6 & 0.626 & \textbf{0.245} & 0.451 & 1.509 & 0.127 & 1.4 \\
        PETR (\citeyear{liu2022petr})                       & C & R101$^*$ & 45.5 & 39.1 & 0.647 & 0.251 & \textbf{0.433} & 0.933 & 0.143 & 1.7 \\
        BEVFormer-S (\citeyear{li2022bevformer})            & C & R101$^*$ & 46.2 & 40.9 & 0.650 & 0.261 & 0.439 & 0.925 & 0.147 & - \\
        \hline
        CenterFusion$^\dagger$ (\citeyear{nabati2021centerfusion}) & C+R & DLA34 & 44.9 & 32.6 & 0.631 & 0.261 & 0.516 & 0.614 & 0.115 & - \\
        CRAFT$^\dagger$                                       & C+R & DLA34 & \textbf{52.3} & \textbf{41.1} & \textbf{0.467} & 0.268 & 0.456 & \textbf{0.519} & \textbf{0.114} & \textbf{4.1} \\
        \hline
    \end{tabular}}
    \end{center}
    \label{table:ns test set}
\end{table*}

\subsection{Detection Heads and Training Objectives}\label{chap:loss}
The fusion detection head decodes the above radar-encoded image proposal feature to refine the image proposal localization and other attributes in polar coordinate.
Specifically, category-specific regression heads on the top of shared MLP layers predict (a) fusion score, (b) location offsets, (c) center-ness, and (d) velocity.

\noindent
(a) Fusion score: Due to radar sparsity, some image proposals are associated with only clutter radar points, which can degrade performance.
Thus we predict the probability of radar points associated with SPA containing at least one radar point from the object.

\noindent
(b) Location offsets: Given the image proposal center, we predict offsets in polar coordinate instead of Cartesian.
We empirically find that this simple and easy to implement technique effectively reduces the significant localization error disagreement in polar coordinates.

\noindent
(c) Center-ness: Following~\cite{yang20203dssd}, we assign higher center-ness scores to predictions closer to ground-truth centers.

\noindent
(d) Velocity: To mitigate the absence of azimuthal velocity in radar Doppler measurement~\cite{long2021full}, we predict the speed and then transform it to velocity using object orientation.

To train the model, we assign the same ground-truth used for training the keypoint-based camera 3D object detector~\cite{law2018cornernet} to image proposal; thus it can be directly matched to the ground truth without an additional matching algorithm~\cite{carion2020end}.
Our final loss function can be formulated as a weighted sum of aforementioned classification and regression losses.
The full loss function is detailed in Appendix \ref{chap:appendix B}.


\setlength{\tabcolsep}{0.3em}
\setcounter{table}{1}
\begin{table*}[!t]
    \caption{Per-class AP comparison on nuScenes \texttt{val} set. 
    `C.V.', `M.C.', and `T.C.' denote construction vehicle, motorcycle, and traffic cone, respectively.}
    \begin{center}
    \resizebox{0.98\textwidth}{!}{
    \begin{tabular}{c|c||cccccccccc|c}
        \hline
        Method & Input &Car&Truck&Bus&Trailer&C.V.&Ped.	&M.C.&Bicycle&T.C.&Barrier&mAP\\
        \hline
        PointPillars (\citeyear{lang2019pointpillars}) & L & 79.9&35.7&42.8&26.1&5.5&71.7&39.4&10.6&33.4&52.0&39.7\\
        FCOS3D (\citeyear{wang2021fcos3d}) & C & 47.9&23.3&31.4&11.2&5.7&41.1&30.5&30.2&55.0&45.5&32.2\\
        \hline
        CenterNet (\citeyear{Zhou2019}) & C & 48.4&23.1&34.0&13.1&3.5&37.7&24.9&23.4&55.0&45.6&30.6\\
        CenterFusion (\citeyear{nabati2021centerfusion}) & C+R & 52.4\scriptsize{(+4.0)}&26.5\scriptsize{(+3.4)}&36.2\scriptsize{(+2.2)}&15.4\scriptsize{(+2.3)}&5.5\scriptsize{(+2.0)}&
        38.9\scriptsize{(+1.2)}&30.5\scriptsize{(+5.6)}&22.9\scriptsize{(-0.5)}&56.3\scriptsize{(+1.3)}&47.0\scriptsize{(+1.4)}&33.2\scriptsize{(+2.6)}\\
        \hline
        CRAFT-I & C & 52.4&25.7&30.0&15.8&5.4&39.3&28.6&29.8&57.5&47.8&33.2\\
        CRAFT & C+R & 
        69.6\scriptsize{(\textbf{+17.2})}&37.6\scriptsize{(\textbf{+11.9})}&47.3\scriptsize{(\textbf{+17.3})}&20.1\scriptsize{(+4.3)}&10.7\scriptsize{(+5.3)}&
        46.2\scriptsize{(+6.9)}&39.5\scriptsize{(\textbf{+10.9})}&31.0\scriptsize{(+1.2)}&57.1\scriptsize{(-0.4)}&51.1\scriptsize{(+3.3)}&41.1\scriptsize{(+7.9)}\\
        \hline
    \end{tabular}}
    \end{center}
    \label{table:ns val set}
\end{table*}

\section{Experiments}
\subsubsection{Dataset and Metrics}
We evaluate our method on a large-scale and challenging nuScenes dataset~\cite{Caesar2020}, which consists of 700/150/150 scenes for \texttt{train}/\texttt{val}/\texttt{test} set.
Each 20 second long sequence has 3D bounding box annotations of 10 classes by 2Hz frequency, and it contains six surrounding camera images, one LiDAR point cloud, and five radar point clouds covering 360 degrees.
The official evaluation metrics are mean Average Precision (mAP), True Positive metrics (\textit{i.e.}, translation, scale, orientation, velocity, and attribute error), and the nuScenes detection score (NDS).
Particularly, NDS is a weighted sum of mAP and True Positive metrics.
The matching thresholds for calculating AP are the center distance of 0.5$m$, 1$m$, 2$m$, and 4$m$, instead of IoU.

\subsubsection{Implementation Details}
We implement our fusion network using CenterNet~\cite{Zhou2019} with DLA34~\cite{Yu2018} backbone.
Note that our camera 3D object detector is trained from scratch on nuScenes without large-scale depth pre-training on DDAD15M proposed in DD3D~\cite{park2021pseudo}.
For the image backbone, we use $448\times800$ size image as network input and use a single feature map of the last layer for the image proposal feature.
For radar, we accumulate six radar sweeps following GRIF Net~\cite{kim2020grif} and set the maximum number of radar points as 2048.
To maximize the recall of the camera 3D detector, we set the image proposal threshold to 0.05 and apply Non-Maximum Suppression (NMS) after fusion.
More detailed implementation details are in Appendix \ref{chap:appendix B}.

\subsubsection{Training and Inference}
As a proof of concept that the proposed method is flexible, we pre-train the camera 3D object detector and keep its weights frozen during training the camera-radar fusion network, while other fusion modules are trained in an end-to-end manner.
Our model takes six surrounding images and five radar point clouds as a single batch to perform 360-degree detection, and data augmentation is applied to both image and radar.
We train our models for 24 epochs with a batch size of 32, cosine annealing scheduler, and $2\times 10^{-4}$ learning rate on 4 RTX 3090 GPUs.
Inference time is measured on an Intel Core i9 CPU and an RTX 3090 GPU without test time augmentation for fusion.
See Appendix \ref{chap:appendix C} for more detailed experimantal settings.

\setlength{\tabcolsep}{0.5em}
\setcounter{table}{2}
\begin{table*}[!t]
    \caption{Ablation experiments to validate our fusion framework design choices.}
    \begin{center}
    \resizebox{0.85\textwidth}{!}{
    \begin{tabular}{c||ccc}
        \multicolumn{4}{c}{(a) Fusion method} \\
        \hline
         & \multicolumn{2}{c}{AP$\uparrow$} &\multirow{2}{*}{ATE$\downarrow$} \\
         & 0.5$m$ & mean & \\
        \hline
        Img.  & 19.6 & 52.4 & 0.49 \\
        FP    & 34.7 & 61.8 & 0.37 \\
        FC    & 38.9 & 63.1 & 0.34 \\
        \hline
        FT  & 41.3 & 65.1 & 0.32 \\
        \hline
    \end{tabular}
    
    \hspace{0.2cm}
    
    \begin{tabular}{c||ccc|c}
        \multicolumn{5}{c}{(b) Association method} \\
        \hline
         & \multicolumn{2}{c}{AP$\uparrow$} &\multirow{2}{*}{ATE$\downarrow$}  &\multirowcell{2}{Assoc.\\RC} \\
         & 0.5$m$ & mean & & \\
        \hline
        No Assoc.  & \multicolumn{3}{c|}{OOM} & 83.5 \\
        RoIPool    & 29.8 & 58.7 & 0.39 & 50.7 \\
        Ball Query & 39.5 & 64.3 & 0.33 & 78.2 \\
        \hline
        SPA        & 41.3 & 65.1 & 0.32 & 77.1 \\
        \hline
    \end{tabular}
    
    \hspace{0.2cm}
    
    \begin{tabular}{c||cc|cc}
        \multicolumn{4}{c}{(c) Coordinate system} \\
        \hline
         & \multicolumn{2}{c|}{AP$\uparrow$} & \multirowcell{2}{Rad.\\Err.} & \multirowcell{2}{Azim.\\Err.} \\
         & 0.5$m$ & mean & & \\
        \hline
        Cart.  &  27.7 & 58.6 & 1.73 & 0.52 \\
        Polar  &  41.3 & 65.1 & 1.26 & 0.25 \\
        \hline
        Imp.   & +13.6 & +6.5 &-0.47 &-0.27 \\
        \hline
    \end{tabular}
    }
    \end{center}
    \label{table:ablation}
\end{table*}

\subsection{Comparison with State-of-the-Arts}
We compare our method with state-of-the-arts on nuScenes \texttt{test} set.
To eliminate the effect of multi-sequence input and large-scale depth pre-training and compare with previous methods fairly, we report methods with single frame input trained on nuScenes dataset only.
Additional comparisons with other methods and results on \texttt{val} set are provided in Appendix \ref{chap:appendix D}. 

As shown in Table \ref{table:ns test set}, our CRAFT outperforms all competing camera-radar and single-frame camera methods on \texttt{test} set.
Importantly, our CRAFT has more than two times faster inference speed compared to other methods.
The most performance gain comes from the improved localization and velocity estimation, which improve the recall at strict thresholds (\textit{e.g.}, 0.5$m$ or 1$m$) and NDS scores, respectively.
As our lightweight CenterNet-like camera 3D detector, denote as CRAFT-I, has no bells and whistles as a proof of concept, we leave a more advanced camera 3D detector for camera-radar fusion as a future work.

Table \ref{table:ns val set} shows the per-class mAP comparison with LiDAR, camera, and camera-radar methods on \texttt{val} set.
Although CenterFusion~\cite{nabati2021centerfusion} and ours have a similar camera 3D detector architecture, our CRAFT achieves a remarkable performance boost (\textbf{+17.2}\% vs. 4.0\% on Car, \textbf{+17.3}\% vs. 2.2\% on Bus), which shows the importance of an appropriate fusion strategy considering sensor characteristics. 
The performance improvement is higher in metallic objects (car, truck, bus and motorcycle) than non-metallic objects (pedestrian, bicycle, traffic cone, and barrier) since metallic objects have more valid radar returns and are thus easier to be distinguished from background clutters.
Trailer and construction vehicle are metallic but have less performance gain, assuming these classes are commonly surrounded by other trailers or construction materials, which can possibly harm the radar performance.
More experiments of robustness against the number of radar points are provided in Appendix \ref{chap:appendix D}.

\subsection{Ablation Studies}
We conduct a series of ablation studies to validate the design of CRAFT on nuScenes \texttt{val} set.
We report the results of the car class for ablation experiments since the performance gains of other classes are less consistent and easier to be fluctuated by other factors.

\subsubsection{Fusion Method}
On top of our CRAFT-I, we conduct additional fusion methods:
using only radar points similar to FPointNet (\textbf{FP})~\cite{qi2018frustum} and fusing features by concatenation (\textbf{FC}) after aggregating radar features by max-pooling.
Table \ref{table:ablation}a shows that our fusion transformer (\textbf{FT}) brings large AP improvement over radar-only (FP, +6.6\%) and naive concatenation (FC, +2.4\%), especially at the strict 0.5$m$ threshold.
It shows that using both image feature and attention-based fusion is beneficial for improving recall and reducing localization error.

\setcounter{figure}{3}
\begin{figure}[t]
\begin{center}
\includegraphics[width=0.94\columnwidth]{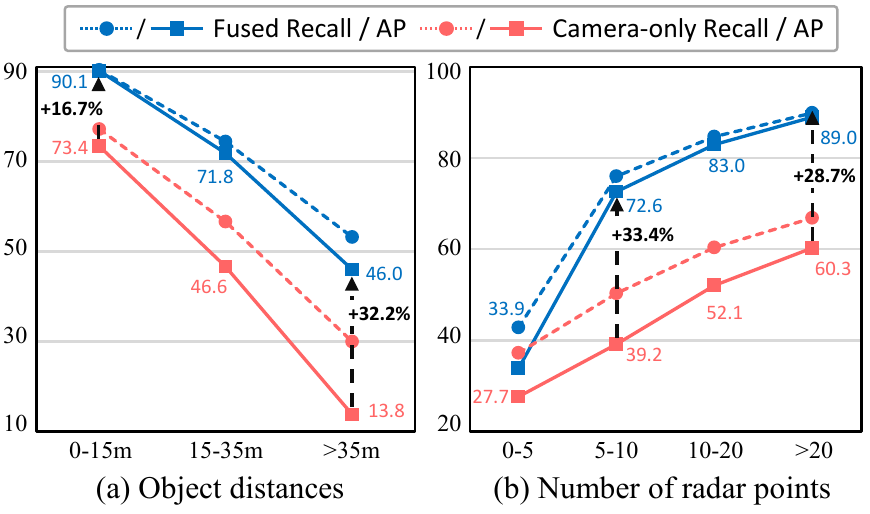}
\end{center}
\caption{
Analysis of different object distances and the number of radar points.
}
\label{fig:analysis}
\end{figure}

\setcounter{figure}{4}
\begin{figure*}[t]
\begin{center}
\includegraphics[width=0.95\textwidth]{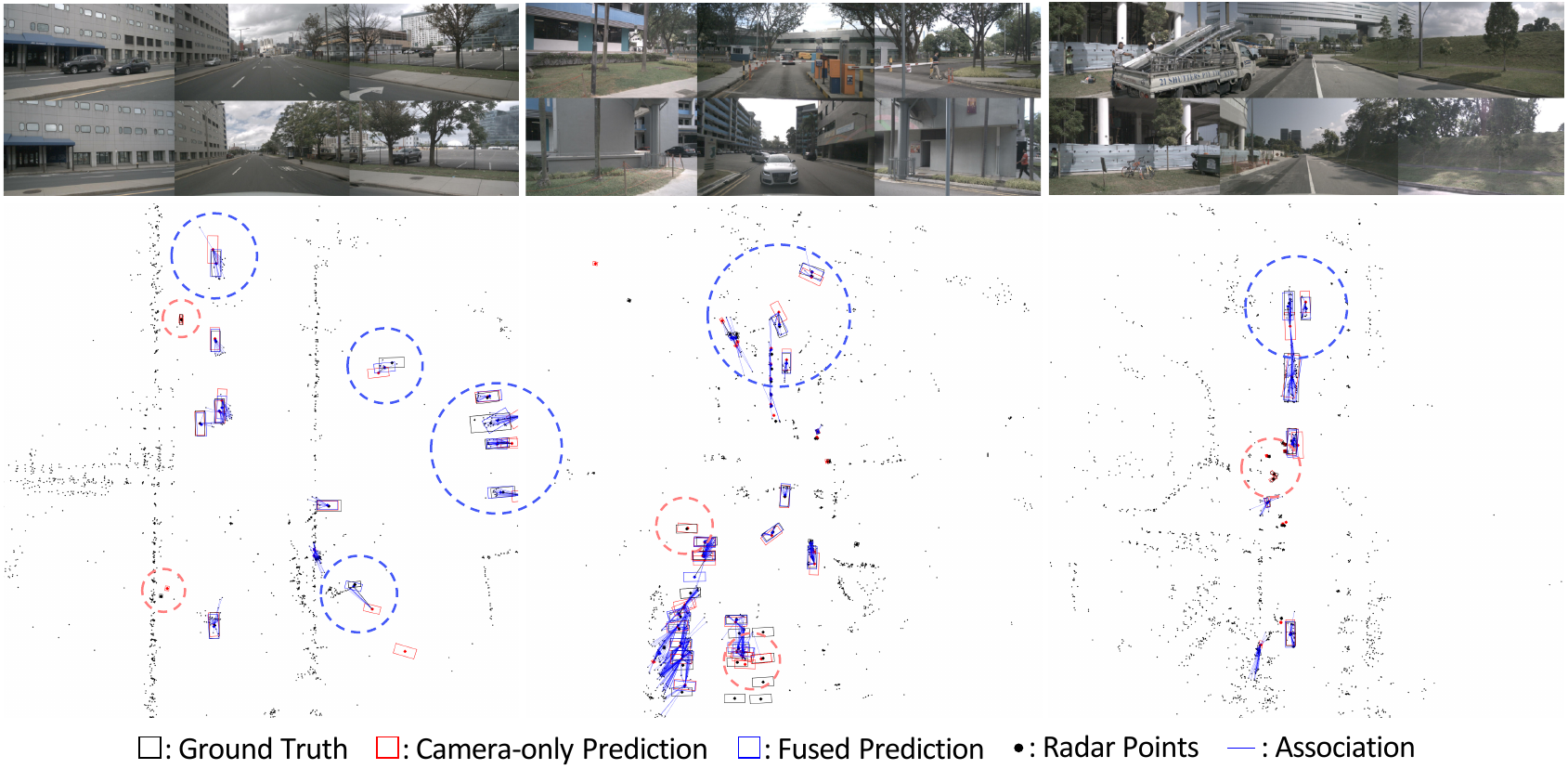}
\end{center}
\caption{Qualitative results of CRAFT.
Blue circles indicate samples that are refined by fusing radar points and have more accurate localization, and red circles indicate samples that are predicted by camera-only since there are no valid radar returns among associated points.
Best viewed in color with zoom in.}
\label{fig:result}
\end{figure*}

\subsubsection{Association Method}
In Table \ref{table:ablation}b, we ablate the proposed association strategy by replacing it with RoIPool (query points inside image proposal) and Ball Query (query points around image proposal center within a 6$m$ radius) to study how the soft polar association benefits the following fusion method.
Association recall denotes the percentage of a proposal containing at least one valid point over associated points.
Taking all points to Radar-to-Image Feature Encoder without association suffers from a substantial computational burden, leading to an out-of-memory in our setting.
RoIPool often fails to associate valid radar points due to inaccurate image proposal localization, and large size ball query contains many clutters that can degrade the fusion performance.
Our SPA enables the best trade-off between recall and precision for effective camera-radar feature fusion. 

\subsubsection{Coordinate System}
Table \ref{table:ablation}c ablates the coordinate system of point association and regression target.
For the Cartesian setting, we associate points using SPA and transform them back to the Cartesian coordinate before feeding them to R2I FE, and also regress offsets in the same coordinate.
Using polar coordinate is similar to applying Principal Component Analysis (PCA) since the error variance of image proposals in the azimuthal direction is minimized in polar coordinate.
Thus, the simple coordinate transformation makes the network learn spatial information easier and effectively reduces the localization error of image proposal.

\subsection{Analysis}
We analyze maximum recall and AP performance under different object distances and the number of radar points using a 1$m$ matching threshold.

\subsubsection{Performance by Distances}
Our camera-radar fusion achieves significant improvements over the camera-only baseline and brings larger improvements for distant objects, as shown in Fig. \ref{fig:analysis}a.
As the depth estimation becomes more inaccurate on distant objects farther than 35$m$, our fusion method benefits more from the radar measurement and achieves a significant improvement of 32.2\%.

\subsubsection{Performance by Radar Points}
We demonstrate the performance under a different number of valid radar points returned from the object in Fig. \ref{fig:analysis}b.
CRAFT achieves consistent improvements over a various number of radar points, which shows that our fusion method is robust to the radar sparsity.
Our method yields robust performance although radar points are not available or only a few (+6.2\%) and shows better performance improvement when more radar points are provided (+28.7\%).

\subsection{Qualitative Results}
We show the detection results in a complex scene in Fig. \ref{fig:result}.
As highlighted with blue circles, CRAFT precisely refines image proposals by adaptively fusing valid radar returns if they are available.
More qualitative results are provided in Appendix \ref{chap:appendix E}.

\section{Conclusions}
In this paper, we have proposed an effective and robust camera-radar 3D detection framework.
With a soft polar association and spatio-contextual fusion transformer, the spatial and contextual information of the camera and radar can be effectively complemented to yield a more accurate prediction given an inaccurate image proposal.
CRAFT achieves state-of-the-art performance on nuScenes by outperforming previous camera-radar and camera-only methods and shows the potential of camera-radar fusion.
Extensive experiments validate the design choice of our framework and show the advantages over a camera-only baseline. 
We hope our work will inspire future research on camera-radar fusion for 3D scene understanding.

\section*{Acknowledgments}
This work was supported by Institute of Information \& communications Technology Planning \& Evaluation (IITP) grant funded by the Korea government (MSIT) (No.2021-0-00951, Development of Cloud based Autonomous Driving AI learning Software).

\bibliography{aaai23}

\begin{thebibliography}{58}
\providecommand{\natexlab}[1]{#1}

\bibitem[{Ba, Kiros, and Hinton(2016)}]{ba2016layer}
Ba, J.~L.; Kiros, J.~R.; and Hinton, G.~E. 2016.
\newblock Layer normalization.
\newblock In \emph{arXiv preprint arXiv:1607.06450}.

\bibitem[{Bai et~al.(2022)Bai, Hu, Zhu, Huang, Chen, Fu, and
  Tai}]{bai2022transfusion}
Bai, X.; Hu, Z.; Zhu, X.; Huang, Q.; Chen, Y.; Fu, H.; and Tai, C.-L. 2022.
\newblock Transfusion: Robust lidar-camera fusion for 3d object detection with
  transformers.
\newblock In \emph{Proceedings of the IEEE/CVF Conference on Computer Vision
  and Pattern Recognition (CVPR)}, 1090--1099.

\bibitem[{Caesar et~al.(2020)Caesar, Bankiti, Lang, Vora, Liong, Xu, Krishnan,
  Pan, Baldan, and Beijbom}]{Caesar2020}
Caesar, H.; Bankiti, V.; Lang, A.~H.; Vora, S.; Liong, V.~E.; Xu, Q.; Krishnan,
  A.; Pan, Y.; Baldan, G.; and Beijbom, O. 2020.
\newblock nuScenes: A multimodal dataset for autonomous driving.
\newblock In \emph{Proceedings of the IEEE/CVF Conference on Computer Vision
  and Pattern Recognition (CVPR)}, 11621--11631.

\bibitem[{Carion et~al.(2020)Carion, Massa, Synnaeve, Usunier, Kirillov, and
  Zagoruyko}]{carion2020end}
Carion, N.; Massa, F.; Synnaeve, G.; Usunier, N.; Kirillov, A.; and Zagoruyko,
  S. 2020.
\newblock End-to-end object detection with transformers.
\newblock In \emph{Proceedings of the European Conference on Computer Vision
  (ECCV)}, 213--229.

\bibitem[{Chen et~al.(2022)Chen, Zhang, Wang, Wang, and Zhao}]{chen2022futr3d}
Chen, X.; Zhang, T.; Wang, Y.; Wang, Y.; and Zhao, H. 2022.
\newblock Futr3d: A unified sensor fusion framework for 3d detection.
\newblock In \emph{Proceedings of the IEEE/CVF Conference on Computer Vision
  and Pattern Recognition (CVPR)}.

\bibitem[{Fu et~al.(2018)Fu, Gong, Wang, Batmanghelich, and Tao}]{Fu2018}
Fu, H.; Gong, M.; Wang, C.; Batmanghelich, K.; and Tao, D. 2018.
\newblock Deep Ordinal Regression Network for Monocular Depth Estimation.
\newblock In \emph{Proceedings of the IEEE/CVF Conference on Computer Vision
  and Pattern Recognition (CVPR)}, 2002--2011.

\bibitem[{Huang et~al.(2020)Huang, Liu, Chen, and Bai}]{huang2020epnet}
Huang, T.; Liu, Z.; Chen, X.; and Bai, X. 2020.
\newblock Epnet: Enhancing point features with image semantics for 3d object
  detection.
\newblock In \emph{Proceedings of the European Conference on Computer Vision
  (ECCV)}, 35--52.

\bibitem[{Hung et~al.(2022)Hung, Kretzschmar, Casser, Hwang, and
  Anguelov}]{hung2022let}
Hung, W.-C.; Kretzschmar, H.; Casser, V.; Hwang, J.-J.; and Anguelov, D. 2022.
\newblock Let-3d-ap: Longitudinal error tolerant 3d average precision for
  camera-only 3d detection.
\newblock In \emph{arXiv preprint arXiv:2206.07705}.

\bibitem[{Ioffe and Szegedy(2015)}]{ioffe2015batch}
Ioffe, S.; and Szegedy, C. 2015.
\newblock Batch normalization: Accelerating deep network training by reducing
  internal covariate shift.
\newblock In \emph{Proceedings of the International Conference on Machine
  Learning (ICML)}, 448--456.

\bibitem[{Johnson and Dudgeon(1992)}]{johnson1992array}
Johnson, D.~H.; and Dudgeon, D.~E. 1992.
\newblock \emph{Array signal processing: concepts and techniques}.
\newblock Simon \& Schuster, Inc.

\bibitem[{Kendall and Gal(2017)}]{kendall2017}
Kendall, A.; and Gal, Y. 2017.
\newblock What Uncertainties Do We Need in Bayesian Deep Learning for Computer
  Vision?
\newblock In \emph{Advances in Neural Information Processing Systems
  (NeurIPS)}, 5574--5584.

\bibitem[{Kim, Kim, and Kum(2020)}]{kim2020low}
Kim, J.; Kim, Y.; and Kum, D. 2020.
\newblock Low-level sensor fusion network for 3D vehicle detection using radar
  range-azimuth heatmap and monocular image.
\newblock In \emph{Proceedings of the Asian Conference on Computer Vision
  (ACCV)}, 388--402.

\bibitem[{Kim, Choi, and Kum(2020)}]{kim2020grif}
Kim, Y.; Choi, J.~W.; and Kum, D. 2020.
\newblock GRIF Net: Gated region of interest fusion network for robust 3D
  object detection from radar point cloud and monocular image.
\newblock In \emph{Proceedings of the IEEE/RSJ International Conference on
  Intelligent Robots and Systems (IROS)}, 10857--10864.

\bibitem[{Kim et~al.(2022)Kim, Kim, Sim, Choi, and Kum}]{kim2022boosting}
Kim, Y.; Kim, S.; Sim, S.; Choi, J.~W.; and Kum, D. 2022.
\newblock Boosting Monocular 3D Object Detection with Object-Centric Auxiliary
  Depth Supervision.
\newblock In \emph{arXiv preprint arXiv:2210.16574}.

\bibitem[{Ku et~al.(2018)Ku, Mozifian, Lee, Harakeh, and
  Waslander}]{ku2018joint}
Ku, J.; Mozifian, M.; Lee, J.; Harakeh, A.; and Waslander, S.~L. 2018.
\newblock Joint 3d proposal generation and object detection from view
  aggregation.
\newblock In \emph{Proceedings of the IEEE/RSJ International Conference on
  Intelligent Robots and Systems (IROS)}, 5750--5757.

\bibitem[{Lang et~al.(2019)Lang, Vora, Caesar, Zhou, Yang, and
  Beijbom}]{lang2019pointpillars}
Lang, A.~H.; Vora, S.; Caesar, H.; Zhou, L.; Yang, J.; and Beijbom, O. 2019.
\newblock Pointpillars: Fast encoders for object detection from point clouds.
\newblock In \emph{Proceedings of the IEEE/CVF Conference on Computer Vision
  and Pattern Recognition (CVPR)}, 12697--12705.

\bibitem[{Law and Deng(2018)}]{law2018cornernet}
Law, H.; and Deng, J. 2018.
\newblock Cornernet: Detecting objects as paired keypoints.
\newblock In \emph{Proceedings of the European Conference on Computer Vision
  (ECCV)}, 734--750.

\bibitem[{Li et~al.(2022)Li, Wang, Li, Xie, Sima, Lu, Yu, and
  Dai}]{li2022bevformer}
Li, Z.; Wang, W.; Li, H.; Xie, E.; Sima, C.; Lu, T.; Yu, Q.; and Dai, J. 2022.
\newblock BEVFormer: Learning Bird's-Eye-View Representation from Multi-Camera
  Images via Spatiotemporal Transformers.
\newblock In \emph{Proceedings of the European Conference on Computer Vision
  (ECCV)}.

\bibitem[{Lim et~al.(2019)Lim, Ansari, Major, Fontijne, Hamilton, Gowaikar, and
  Subramanian}]{lim2019radar}
Lim, T.-Y.; Ansari, A.; Major, B.; Fontijne, D.; Hamilton, M.; Gowaikar, R.;
  and Subramanian, S. 2019.
\newblock Radar and camera early fusion for vehicle detection in advanced
  driver assistance systems.
\newblock In \emph{Advances in Neural Information Processing Systems Workshops
  (NeurIPSW)}.

\bibitem[{Lin, Dai, and Van~Gool(2020)}]{lin2020depth}
Lin, J.-T.; Dai, D.; and Van~Gool, L. 2020.
\newblock Depth estimation from monocular images and sparse radar data.
\newblock In \emph{Proceedings of the IEEE/RSJ International Conference on
  Intelligent Robots and Systems (IROS)}, 10233--10240.

\bibitem[{Lin et~al.(2018)Lin, Le~Kernec, Yang, Fioranelli, Romain, and
  Zhao}]{lin2018human}
Lin, Y.; Le~Kernec, J.; Yang, S.; Fioranelli, F.; Romain, O.; and Zhao, Z.
  2018.
\newblock Human activity classification with radar: Optimization and noise
  robustness with iterative convolutional neural networks followed with random
  forests.
\newblock \emph{IEEE Sensors Journal}, 18(23): 9669--9681.

\bibitem[{Liu et~al.(2022{\natexlab{a}})Liu, Wang, Zhang, and
  Sun}]{liu2022petr}
Liu, Y.; Wang, T.; Zhang, X.; and Sun, J. 2022{\natexlab{a}}.
\newblock PETR: Position Embedding Transformation for Multi-View 3D Object
  Detection.
\newblock In \emph{Proceedings of the European Conference on Computer Vision
  (ECCV)}, 531--–548.

\bibitem[{Liu et~al.(2022{\natexlab{b}})Liu, Tang, Amini, Yang, Mao, Rus, and
  Han}]{liu2022bevfusion}
Liu, Z.; Tang, H.; Amini, A.; Yang, X.; Mao, H.; Rus, D.; and Han, S.
  2022{\natexlab{b}}.
\newblock BEVFusion: Multi-Task Multi-Sensor Fusion with Unified Bird's-Eye
  View Representation.
\newblock In \emph{Advances in Neural Information Processing Systems
  (NeurIPS)}.

\bibitem[{Long et~al.(2021{\natexlab{a}})Long, Morris, Liu, Castro,
  Chakravarty, and Narayanan}]{long2021full}
Long, Y.; Morris, D.; Liu, X.; Castro, M.; Chakravarty, P.; and Narayanan, P.
  2021{\natexlab{a}}.
\newblock Full-Velocity Radar Returns by Radar-Camera Fusion.
\newblock In \emph{Proceedings of the IEEE/CVF International Conference on
  Computer Vision (ICCV)}, 16198--16207.

\bibitem[{Long et~al.(2021{\natexlab{b}})Long, Morris, Liu, Castro,
  Chakravarty, and Narayanan}]{long2021radar}
Long, Y.; Morris, D.; Liu, X.; Castro, M.; Chakravarty, P.; and Narayanan, P.
  2021{\natexlab{b}}.
\newblock Radar-camera pixel depth association for depth completion.
\newblock In \emph{Proceedings of the IEEE/CVF Conference on Computer Vision
  and Pattern Recognition (CVPR)}, 12507--12516.

\bibitem[{Loshchilov and Hutter(2019)}]{loshchilov2018decoupled}
Loshchilov, I.; and Hutter, F. 2019.
\newblock Decoupled Weight Decay Regularization.
\newblock In \emph{Proceedings of the International Conference on Learning
  Representations (ICLR)}.

\bibitem[{Ma et~al.(2021)Ma, Zhang, Xu, Zhou, Yi, Li, and
  Ouyang}]{ma2021delving}
Ma, X.; Zhang, Y.; Xu, D.; Zhou, D.; Yi, S.; Li, H.; and Ouyang, W. 2021.
\newblock Delving into Localization Errors for Monocular 3D Object Detection.
\newblock In \emph{Proceedings of the IEEE/CVF Conference on Computer Vision
  and Pattern Recognition (CVPR)}, 4721--4730.

\bibitem[{Major et~al.(2019)Major, Fontijne, Ansari, Teja~Sukhavasi, Gowaikar,
  Hamilton, Lee, Grzechnik, and Subramanian}]{major2019vehicle}
Major, B.; Fontijne, D.; Ansari, A.; Teja~Sukhavasi, R.; Gowaikar, R.;
  Hamilton, M.; Lee, S.; Grzechnik, S.; and Subramanian, S. 2019.
\newblock Vehicle detection with automotive radar using deep learning on
  range-azimuth-doppler tensors.
\newblock In \emph{Proceedings of the IEEE/CVF International Conference on
  Computer Vision Workshops (ICCVW)}, 924--932.

\bibitem[{Meyer and Kuschk(2019{\natexlab{a}})}]{meyer2019automotive}
Meyer, M.; and Kuschk, G. 2019{\natexlab{a}}.
\newblock Automotive radar dataset for deep learning based 3d object detection.
\newblock In \emph{Proceedings of the European Radar Conference (EuRAD)},
  129--132.

\bibitem[{Meyer and Kuschk(2019{\natexlab{b}})}]{meyer2019deep}
Meyer, M.; and Kuschk, G. 2019{\natexlab{b}}.
\newblock Deep learning based 3d object detection for automotive radar and
  camera.
\newblock In \emph{Proceedings of the European Radar Conference (EuRAD)},
  133--136.

\bibitem[{Misra, Girdhar, and Joulin(2021)}]{misra2021end}
Misra, I.; Girdhar, R.; and Joulin, A. 2021.
\newblock An end-to-end transformer model for 3d object detection.
\newblock In \emph{Proceedings of the IEEE/CVF International Conference on
  Computer Vision (ICCV)}, 2906--2917.

\bibitem[{Nabati and Qi(2021)}]{nabati2021centerfusion}
Nabati, R.; and Qi, H. 2021.
\newblock Centerfusion: Center-based radar and camera fusion for 3d object
  detection.
\newblock In \emph{Proceedings of the IEEE/CVF Winter Conference on
  Applications of Computer Vision (WACV)}, 1527--1536.

\bibitem[{Pan et~al.(2021)Pan, Xia, Song, Li, and Huang}]{pan20213d}
Pan, X.; Xia, Z.; Song, S.; Li, L.~E.; and Huang, G. 2021.
\newblock 3d object detection with pointformer.
\newblock In \emph{Proceedings of the IEEE/CVF Conference on Computer Vision
  and Pattern Recognition (CVPR)}, 7463--7472.

\bibitem[{Park et~al.(2021)Park, Ambrus, Guizilini, Li, and
  Gaidon}]{park2021pseudo}
Park, D.; Ambrus, R.; Guizilini, V.; Li, J.; and Gaidon, A. 2021.
\newblock Is Pseudo-Lidar needed for Monocular 3D Object detection?
\newblock In \emph{Proceedings of the IEEE/CVF International Conference on
  Computer Vision (ICCV)}, 3142--3152.

\bibitem[{Philion and Fidler(2020)}]{philion2020lift}
Philion, J.; and Fidler, S. 2020.
\newblock Lift, splat, shoot: Encoding images from arbitrary camera rigs by
  implicitly unprojecting to 3d.
\newblock In \emph{Proceedings of the European Conference on Computer Vision
  (ECCV)}, 194--210.

\bibitem[{Qi et~al.(2018)Qi, Liu, Wu, Su, and Guibas}]{qi2018frustum}
Qi, C.~R.; Liu, W.; Wu, C.; Su, H.; and Guibas, L.~J. 2018.
\newblock Frustum pointnets for 3d object detection from rgb-d data.
\newblock In \emph{Proceedings of the IEEE/CVF Conference on Computer Vision
  and Pattern Recognition (CVPR)}, 918--927.

\bibitem[{Qi et~al.(2017{\natexlab{a}})Qi, Su, Mo, and Guibas}]{qi2017pointnet}
Qi, C.~R.; Su, H.; Mo, K.; and Guibas, L.~J. 2017{\natexlab{a}}.
\newblock Pointnet: Deep learning on point sets for 3d classification and
  segmentation.
\newblock In \emph{Proceedings of the IEEE/CVF Conference on Computer Vision
  and Pattern Recognition (CVPR)}, 652--660.

\bibitem[{Qi et~al.(2017{\natexlab{b}})Qi, Yi, Su, and
  Guibas}]{qi2017pointnet++}
Qi, C.~R.; Yi, L.; Su, H.; and Guibas, L.~J. 2017{\natexlab{b}}.
\newblock Pointnet++: Deep hierarchical feature learning on point sets in a
  metric space.
\newblock In \emph{Advances in Neural Information Processing Systems
  (NeurIPS)}, 5105--5114.

\bibitem[{Reading et~al.(2021)Reading, Harakeh, Chae, and
  Waslander}]{reading2021categorical}
Reading, C.; Harakeh, A.; Chae, J.; and Waslander, S.~L. 2021.
\newblock Categorical depth distribution network for monocular 3d object
  detection.
\newblock In \emph{Proceedings of the IEEE/CVF Conference on Computer Vision
  and Pattern Recognition (CVPR)}, 8555--8564.

\bibitem[{Ren et~al.(2015)Ren, He, Girshick, and Sun}]{Sun2015}
Ren, S.; He, K.; Girshick, R.; and Sun, J. 2015.
\newblock {Faster R-CNN: Towards Real-Time Object Detection with Region
  Proposal Networks}.
\newblock In \emph{Advances in Neural Information Processing Systems
  (NeurIPS)}, 91--99.

\bibitem[{Shi, Wang, and Li(2019)}]{shi2019pointrcnn}
Shi, S.; Wang, X.; and Li, H. 2019.
\newblock Pointrcnn: 3d object proposal generation and detection from point
  cloud.
\newblock In \emph{Proceedings of the IEEE/CVF Conference on Computer Vision
  and Pattern Recognition (CVPR)}, 770--779.

\bibitem[{Svenningsson, Fioranelli, and Yarovoy(2021)}]{svenningsson2021radar}
Svenningsson, P.; Fioranelli, F.; and Yarovoy, A. 2021.
\newblock Radar-pointgnn: Graph based object recognition for unstructured radar
  point-cloud data.
\newblock In \emph{Proceedings of the IEEE Radar Conference (RadarConf)}, 1--6.

\bibitem[{Thomas et~al.(2019)Thomas, Qi, Deschaud, Marcotegui, Goulette, and
  Guibas}]{thomas2019kpconv}
Thomas, H.; Qi, C.~R.; Deschaud, J.-E.; Marcotegui, B.; Goulette, F.; and
  Guibas, L.~J. 2019.
\newblock Kpconv: Flexible and deformable convolution for point clouds.
\newblock In \emph{Proceedings of the IEEE/CVF International Conference on
  Computer Vision (ICCV)}, 6411--6420.

\bibitem[{Tian et~al.(2019)Tian, Shen, Chen, and He}]{tian2019fcos}
Tian, Z.; Shen, C.; Chen, H.; and He, T. 2019.
\newblock Fcos: Fully convolutional one-stage object detection.
\newblock In \emph{Proceedings of the IEEE/CVF International Conference on
  Computer Vision (ICCV)}, 9627--9636.

\bibitem[{Ulrich et~al.(2022)Ulrich, Braun, K{\"o}hler, Niederl{\"o}hner,
  Faion, Gl{\"a}ser, and Blume}]{ulrich2022improved}
Ulrich, M.; Braun, S.; K{\"o}hler, D.; Niederl{\"o}hner, D.; Faion, F.;
  Gl{\"a}ser, C.; and Blume, H. 2022.
\newblock Improved Orientation Estimation and Detection with Hybrid Object
  Detection Networks for Automotive Radar.
\newblock In \emph{arXiv preprint arXiv:2205.02111}.

\bibitem[{Vaswani et~al.(2017)Vaswani, Shazeer, Parmar, Uszkoreit, Jones,
  Gomez, Kaiser, and Polosukhin}]{vaswani2017attention}
Vaswani, A.; Shazeer, N.; Parmar, N.; Uszkoreit, J.; Jones, L.; Gomez, A.~N.;
  Kaiser, {\L}.; and Polosukhin, I. 2017.
\newblock Attention is all you need.
\newblock In \emph{Advances in Neural Information Processing Systems
  (NeurIPS)}, 6000--6010.

\bibitem[{Vora et~al.(2020)Vora, Lang, Helou, and
  Beijbom}]{vora2020pointpainting}
Vora, S.; Lang, A.~H.; Helou, B.; and Beijbom, O. 2020.
\newblock Pointpainting: Sequential fusion for 3d object detection.
\newblock In \emph{Proceedings of the IEEE/CVF Conference on Computer Vision
  and Pattern Recognition (CVPR)}, 4604--4612.

\bibitem[{Wang et~al.(2021{\natexlab{a}})Wang, Xinge, Pang, and
  Lin}]{wang2021probabilistic}
Wang, T.; Xinge, Z.; Pang, J.; and Lin, D. 2021{\natexlab{a}}.
\newblock Probabilistic and geometric depth: Detecting objects in perspective.
\newblock In \emph{Proceedings of the Conference on Robot Learning (CoRL)},
  1475--1485.

\bibitem[{Wang et~al.(2021{\natexlab{b}})Wang, Zhu, Pang, and
  Lin}]{wang2021fcos3d}
Wang, T.; Zhu, X.; Pang, J.; and Lin, D. 2021{\natexlab{b}}.
\newblock Fcos3d: Fully convolutional one-stage monocular 3d object detection.
\newblock In \emph{Proceedings of the IEEE/CVF International Conference on
  Computer Vision Workshops (ICCVW)}, 913--922.

\bibitem[{Wang et~al.(2022)Wang, Guizilini, Zhang, Wang, Zhao, and
  Solomon}]{wang2022detr3d}
Wang, Y.; Guizilini, V.~C.; Zhang, T.; Wang, Y.; Zhao, H.; and Solomon, J.
  2022.
\newblock Detr3d: 3d object detection from multi-view images via 3d-to-2d
  queries.
\newblock In \emph{Proceedings of the Conference on Robot Learning (CoRL)},
  180--191.

\bibitem[{Xiong et~al.(2020)Xiong, Yang, He, Zheng, Zheng, Xing, Zhang, Lan,
  Wang, and Liu}]{xiong2020layer}
Xiong, R.; Yang, Y.; He, D.; Zheng, K.; Zheng, S.; Xing, C.; Zhang, H.; Lan,
  Y.; Wang, L.; and Liu, T. 2020.
\newblock On layer normalization in the transformer architecture.
\newblock In \emph{Proceedings of the International Conference on Machine
  Learning (ICML)}, 10524--10533.

\bibitem[{Yang et~al.(2020)Yang, Sun, Liu, and Jia}]{yang20203dssd}
Yang, Z.; Sun, Y.; Liu, S.; and Jia, J. 2020.
\newblock 3dssd: Point-based 3d single stage object detector.
\newblock In \emph{Proceedings of the IEEE/CVF Conference on Computer Vision
  and Pattern Recognition (CVPR)}, 11040--11048.

\bibitem[{Yin, Zhou, and Krahenbuhl(2021)}]{yin2021center}
Yin, T.; Zhou, X.; and Krahenbuhl, P. 2021.
\newblock Center-based 3d object detection and tracking.
\newblock In \emph{Proceedings of the IEEE/CVF Conference on Computer Vision
  and Pattern Recognition (CVPR)}, 11784--11793.

\bibitem[{Yoo et~al.(2020)Yoo, Kim, Kim, and Choi}]{yoo20203d}
Yoo, J.~H.; Kim, Y.; Kim, J.; and Choi, J.~W. 2020.
\newblock 3d-cvf: Generating joint camera and lidar features using cross-view
  spatial feature fusion for 3d object detection.
\newblock In \emph{Proceedings of the European Conference on Computer Vision
  (ECCV)}, 720--736.

\bibitem[{Yu et~al.(2018)Yu, Wang, Shelhamer, and Darrell}]{Yu2018}
Yu, F.; Wang, D.; Shelhamer, E.; and Darrell, T. 2018.
\newblock {Deep Layer Aggregation}.
\newblock In \emph{Proceedings of the IEEE/CVF Conference on Computer Vision
  and Pattern Recognition (CVPR)}, 2403--2412.

\bibitem[{Zhou, Wang, and Kr{\"a}henb{\"u}hl(2019)}]{Zhou2019}
Zhou, X.; Wang, D.; and Kr{\"a}henb{\"u}hl, P. 2019.
\newblock Objects as Points.
\newblock In \emph{arXiv preprint arXiv:1904.07850}.

\bibitem[{Zhu et~al.(2019)Zhu, Jiang, Zhou, Li, and Yu}]{zhu2019class}
Zhu, B.; Jiang, Z.; Zhou, X.; Li, Z.; and Yu, G. 2019.
\newblock Class-balanced grouping and sampling for point cloud 3d object
  detection.
\newblock In \emph{arXiv preprint arXiv:1908.09492}.

\bibitem[{Zhu et~al.(2021)Zhu, Su, Lu, Li, Wang, and Dai}]{zhu2020deformable}
Zhu, X.; Su, W.; Lu, L.; Li, B.; Wang, X.; and Dai, J. 2021.
\newblock Deformable detr: Deformable transformers for end-to-end object
  detection.
\newblock In \emph{Proceedings of the International Conference on Learning
  Representations (ICLR)}.

\end{thebibliography}

\vspace{1cm}
{\Huge \textbf{Appendix}}

\renewcommand\thesection{\Alph{section}}
\setcounter{section}{0}

\section{Architectural Details} \label{chap:appendix A}
\subsection{Camera 3D Detector}
We adopt CenterNet~\cite{Zhou2019} as a camera-only baseline and make modifications to detection heads, label assignment strategy, and post-processing.
We denote our improved CenterNet as CRAFT-I.

For the detection head, we add the depth variance prediction head and replace the L1 regression loss of CenterNet with uncertainty-aware regression loss following~\cite{kendall2017} as:
\setcounter{equation}{4}
\begin{equation}
\mathcal{L}_{dep}=\frac{1}{N}\sum_{i=1}^{N}\left[\frac{\lVert d_{i}^*-\hat{d}_{i}\rVert_1}{\hat{\sigma}_i^2} + \log \hat{\sigma}_i^2\right],
\label{eq:depth loss}
\end{equation}
where $N$ denotes the number of objects, $d_{i}^*$ and $d_i$ are the ground truth and predicted depth at the object keypoint location.
In practice, we predict the variance $\hat{\sigma}^2$ as log variance $s:=\log\hat{\sigma}^2$ for numerical stability.
The convolutional network for all detection heads consists of $3\times3$ and $1\times1$ convolutional layers with ReLU, which is identical to CenterNet.
We refer the reader to CenterNet~\cite{Zhou2019} and MonoPixel~\cite{kim2022boosting} for more details of architecture.

Unlike CenterNet detects the center of the 2D bounding box as a keypoint and additionally predicts the pixel offset to the projected 3D center, we define the projected 3D center as the keypoint similar to~\cite{ma2021delving}.
Consequently, we assign the ground truth to the projected 3D center and regress other attributes from the keypoint location.

For post-processing, we readjust the confidence of the predicted object using the depth variance.
We first map the depth variance $\sigma^2$ estimated by the network into the confidence form $p_{dep}\in[0,1]^N$ via the negative exponential function and then estimate the 3D confidence as a combination of depth confidence $p_{dep}$ and class confidence $p_{k}$ as:
\begin{equation}
p_{dep}=e^{-\sigma^2}, p_{3D}=p_{dep}p_{k}.
\label{eq:confidence}
\end{equation}
Since the depth confidence has a lower value when the estimated depth has a higher variance, this can better represent the localization probability of the predicted object.


\subsection{Radar Backbone}
The number of radar points significantly varies by the driving environment and points are sparse compared to LiDAR.
Specifically, the average and standard deviation of LiDAR and \textit{six sweeps accumulated} radar points in nuScenes \texttt{val} set are $34317\pm271$ and $1589\pm510$.
We choose point representation instead of voxel or pillar, considering the point-wise backbone can be more efficient than the convolutional backbone when points are sparse.

We adopt PointNet++~\cite{qi2017pointnet++} four layers of set abstraction (SA) and two layers of feature propagation (FP) modules for radar feature extraction.
Subsampling strategy such as farthest point sampling (FPS) is discarded, taking into account the sparsity of radar. 
For the same reason, layer normalization~\cite{ba2016layer} instead of batch normalization~\cite{ioffe2015batch} is used in PointNet. 
The detailed network architecture is provided in Table \ref{table:pointnet details}.
We empirically find that using multi-scale grouping (MSG), more nsample, or deeper layers does not greatly affect the performance in our fusion framework.

\setlength{\tabcolsep}{0.6em}
\setcounter{table}{3}
\begin{table}[!t]
    \caption{Detailed architecture of radar backbone.}
    \begin{center}
    \resizebox{0.98\columnwidth}{!}{
    \begin{tabular}{c|cccc}
        \hline
        layer & npoint & radius & nsample & mlp  \\
        \hline
        SA1  & 2048   & 0.4    & 4       & (5, 16, 16, 32) \\
        SA2  & 2048   & 0.8    & 4       & (32, 32, 32, 64) \\
        SA3  & 2048   & 1.2    & 8       & (64, 64, 128, 128) \\
        SA4  & 2048   & 1.6    & 8       & (128, 128, 256, 256) \\
        \hline
        FP1  & - & - & - & (256+128, 256, 256) \\
        FP2  & - & - & - & (256+64, 128, 64) \\
        \hline
    \end{tabular}}
    \end{center}
    \label{table:pointnet details}
\end{table}


\subsection{Spatio-Contextual Fusion Transformer}
Our spatio-contextual fusion transformer takes image proposal and associated radar points features with 64 channels.
We use $M=64$ for the maximum number of image proposals per image and $K^{'}=128$ for the maximum number of radar associations per image proposal.
For Image-to-Radar Feature Encoder (I2R FE), $7\times7$ size image patch features are first projected to the same dimension with single layer MLP, then fed to the deformable multi-head cross-attention (D-MCA) module.
We use the D-MCA module using a single-scale feature map and four sampling points without specific modification.
Given the output of I2R FE, which is the image context encoded radar point feature, we use 2-layer multi-layer-perceptron (MLP) to predict the per-point probability of being inside the 3D bounding box.
This auxiliary point-wise classification task is trained by binary cross-entropy loss.
For Radar-to-Image Feature Encoder (R2I FE), we omit the z-axis of radar points before feeding them into learned positional embedding and R2I FE since the radar does not provide relevant elevation information.
All Transformer modules have 8 heads for multi-head attention, and MLPs have 2-layer with 256 size hidden dimensions.
We use four layers of I2R FE and R2I FE, respectively.


\setlength{\tabcolsep}{0.6em}
\setcounter{table}{4}
\begin{table*}[!t]
    \caption{
    Additional state-of-the-art comparison.
    `L', `C', and `R' represent LiDAR, camera, and radar, respectively.
    $*$ are initialized with FCOS3D~\cite{wang2021fcos3d} checkpoint and $**$ are pre-trained on the depth estimation task with extra data~\cite{park2021pseudo}.
    $\ddagger$ takes multi-frame images as input and $\dagger$ uses image flipping test time augmentation.
    }
    \begin{center}
    \resizebox{0.95\textwidth}{!}{
    \begin{tabular}{c|c|c|c||cc|ccccc|c}
        \hline
        Split & Method & Input & Backbone & NDS$\uparrow$ & mAP$\uparrow$ & mATE$\downarrow$ & mASE$\downarrow$ & mAOE$\downarrow$ & mAVE$\downarrow$ & mAAE$\downarrow$ & FPS \\
        \hline
        \multirow{20}{*}{\texttt{test}} & PointPillars (\citeyear{lang2019pointpillars})    & L & - & 45.3 & 30.5 & 0.517 & 0.290 & 0.500 & 0.316 & 0.368 & 61 \\
        &CenterPoint (\citeyear{yin2021center})           & L & - & 67.3 & 60.3 & 0.262 & 0.239 & 0.361 & 0.288 & 0.136 & 30 \\
        \cline{2-12}
        &TransFusion (\citeyear{bai2022transfusion})       & C+L & DLA34 & 71.7 & 68.9 & 0.259 & 0.243 & 0.359 & 0.288 & 0.127 & 3.7 \\
        &BEVFusion (\citeyear{liu2022bevfusion})           & C+L & Swin-T & 72.9 & 70.2 & 0.261 & 0.239 & 0.329 & 0.260 & 0.134 & 8.4 \\
        \cline{2-12}
        &Radar-PointGNN (\citeyear{svenningsson2021radar}) & R & - &   3.4 & 0.5 & 1.024 & 0.859 & 0.897 & 1.020 & 0.931 & - \\
        &KPConvPillars (\citeyear{ulrich2022improved})     & R & - &  13.9 & 4.9 & 0.823 & 0.428 & 0.607 & 2.081 & 1.000 & - \\
        \cline{2-12}
        &CenterNet$^\dagger$ (\citeyear{Zhou2019})                     & C & HGLS & 40.0 & 33.8 & 0.658 & 0.255 & 0.629 & 1.629 & 0.142 & - \\
        &FCOS3D$^\dagger$ (\citeyear{wang2021fcos3d})                  & C & R101 & 42.8 & 35.8 & 0.690 & 0.249 & 0.452 & 1.434 & 0.124 & 1.7 \\
        &PGD$^\dagger$ (\citeyear{wang2021probabilistic})              & C & R101 & 44.8 & 38.6 & 0.626 & 0.245 & 0.451 & 1.509 & 0.127 & 1.4 \\
        &PETR (\citeyear{liu2022petr})                       & C & R101$^*$ & 45.5 & 39.1 & 0.647 & 0.251 & 0.433 & 0.933 & 0.143 & 1.7 \\
        &BEVFormer-S (\citeyear{li2022bevformer})            & C & R101$^*$ & 46.2 & 40.9 & 0.650 & 0.261 & 0.439 & 0.925 & 0.147 & - \\
        &PETR (\citeyear{liu2022petr})                       & C & Swin-S   & 48.1 & 43.4 & 0.641 & 0.248 & 0.437 & 0.894 & 0.143 & - \\
        \cline{2-12}
        &DD3D$^\dagger$ (\citeyear{park2021pseudo})            & C & V2-99$^{**}$ & 47.7 & 41.8 & 0.572 & 0.249 & 0.368 & 1.014 & 0.124 & - \\
        &DETR3D (\citeyear{wang2022detr3d})                    & C & V2-99$^{**}$ & 47.9 & 41.2 & 0.641 & 0.255 & 0.394 & 0.845 & 0.133 & - \\
        &BEVFormer-S (\citeyear{li2022bevformer})              & C & V2-99$^{**}$& 49.5 & 43.5 & 0.589 & 0.254 & 0.402 & 0.842 & 0.131 & - \\
        &PETR (\citeyear{liu2022petr})                         & C & V2-99$^{**}$& 50.4 & 44.1 & 0.593 & 0.249 & 0.383 & 0.808 & 0.132 & - \\
        &BEVFormer$^{\ddagger}$ (\citeyear{li2022bevformer})        & C & R101$^{*}$   & 53.5 & 44.5 & 0.631 & 0.257 & 0.405 & 0.435 & 0.143 & 1.7 \\
        &BEVFormer$^{\ddagger}$ (\citeyear{li2022bevformer})        & C & V2-99$^{**}$ & 56.9 & 48.1 & 0.582 & 0.256 & 0.375 & 0.378 & 0.126 & - \\
        \cline{2-12}
        &CenterFusion$^\dagger$ (\citeyear{nabati2021centerfusion}) & C+R & DLA34 & 44.9 & 32.6 & 0.631 & 0.261 & 0.516 & 0.614 & 0.115 & - \\
        &CRAFT$^\dagger$                                       & C+R & DLA34 & 52.3 & 41.1 & 0.467 & 0.268 & 0.456 & 0.519 & 0.114 & 4.1 \\
        \hline
        \multirow{5}{*}{\texttt{val}} 
        &CenterNet$^\dagger$ (\citeyear{Zhou2019})                     & C & DLA34 & 32.8 & 30.6 & 0.716 & 0.264 & 0.609 & 1.426 & 0.658 & - \\
        &CRAFT-I$^\dagger$  &  C  & DLA34 & 40.8 & 32.2 & 0.698 & 0.273 & 0.436 & 1.062 & 0.171 & 4.7 \\
        \cline{2-12}
        &CenterFusion$^\dagger$ (\citeyear{nabati2021centerfusion})  & C+R & DLA34 & 45.3 & 33.2 & 0.649 & 0.263 & 0.535 & 0.540 & 0.142 & - \\
        &FUTR3D (\citeyear{chen2022futr3d})  & C+R & R50 & 45.9 & 35.0 & - & - & - & 0.561 & - & - \\
        &CRAFT$^\dagger$    & C+R & DLA34 & 51.7 & 41.1 & 0.494 & 0.276 & 0.454 & 0.486 & 0.176 & 4.1 \\
        \hline
    \end{tabular}}
    \end{center}
    \label{table:ns additional test set}
\end{table*}

\setlength{\tabcolsep}{0.4em}
\setcounter{table}{5}
\begin{table*}[!t]
    \caption{
    Detailed performance comparison of \textit{Car} class with different distance thresholds.
    }
    \begin{center}
    \resizebox{1.0\textwidth}{!}{
    \begin{tabular}{c|c|c|c||cccc|ccccc}
        \hline
        \multirow{2}{*}{Split} & \multirow{2}{*}{Method} & \multirow{2}{*}{Input} & \multirow{2}{*}{Backbone} & \multicolumn{4}{c|}{AP$\uparrow$ [$\%$]} & ATE$\downarrow$  & ASE$\downarrow$  & AOE$\downarrow$ & AVE$\downarrow$ & AAE$\downarrow$\\
        & & & & 0.5$m$ & 1$m$ & 2$m$ & 4$m$ & [$m$] & [1-IoU] & [rad] & [m/s] & [1-acc] \\
        \hline
        \multirow{10}{*}{\texttt{test}} & PointPillars    & L & - & 53.0 & 69.6 & 74.1 & 76.9 & 0.28 & 0.16 & 0.20 & 0.24 & 0.36 \\
        \cline{2-13}
        &Radar-PointGNN & R & - &  0.0 &  0.0 &  5.0 & 15.8 & 1.11 & 0.20 & 0.72 & 1.16 & 0.45 \\
        &KPConvPillars  & R & - &  6.2 & 24.2 & 39.9 & 48.8 & 0.59 & 0.18 & 0.34 & 2.10 & 1.00 \\
        \cline{2-13}
        &FCOS3D      & C & R101  & 15.3 & 43.8 & 68.9 & 81.7 & 0.56 & 0.15 & 0.09 & 1.60 & 0.11 \\
        &CenterNet   & C & HGLS  & 20.0 & 45.8 & 68.0 & 80.6 & 0.47 & 0.14 & 0.09 & 1.72 & 0.13 \\
        &PGD         & C & R101  & 20.4 & 48.7 & 71.8 & 83.4 & 0.48 & 0.15 & 0.10 & 1.70 & 0.11 \\
        &PETR        & C & Swin-S& 24.6 & 53.9 & 76.7 & 85.9 & 0.47 & 0.15 & 0.07 & 0.70 & 0.14 \\
        &BEVFormer$^{\ddagger}$ & C & R101$^*$  & 29.1 & 58.4 & 78.7 & 87.0 & 0.44 & 0.14 & 0.07 & 0.27 & 0.12 \\
        \cline{2-13}
        &CenterFusion& C+R & DLA34 & 22.3 & 45.7 & 62.6 & 72.9 & 0.44 & 0.14 & 0.09 & 0.41 & 0.11 \\
        &CRAFT       & C+R & DLA34 & 50.4\scriptsize{(\textbf{+28.1})} & 66.9\scriptsize{(\textbf{+21.2})} & 74.7\scriptsize{(\textbf{+12.1})} & 80.5\scriptsize{(+7.6)} & 0.27\scriptsize{(-0.17)} & 0.16\scriptsize{(+0.02)} & 0.07\scriptsize{(-0.02)} & 0.28\scriptsize{(-0.13)} & 0.13\scriptsize{(+0.02)} \\
        \hline
        \multirow{2}{*}{\texttt{val}} & CRAFT-I & C & DLA34 & 19.6 & 45.4 & 66.8 & 77.8 & 0.49 & 0.16 & 0.07 & 1.16 & 0.15 \\
        &CRAFT   &C+R& DLA34 & 51.9\scriptsize{(\textbf{+32.3})} & 68.8\scriptsize{(\textbf{+23.4})} & 76.2\scriptsize{(+9.4)} & 81.6\scriptsize{(+3.8)} & 0.26\scriptsize{(-0.23)} & 0.16 & 0.06\scriptsize{(-0.01)} & 0.31\scriptsize{(-0.85)} & 0.15 \\
        \hline
    \end{tabular}}
    \end{center}
    \label{table:additional per distance results}
\end{table*}

\section{Implementation Details} \label{chap:appendix B}
\subsection{Pre-processing and Hyper-parameters}
The camera 3D detection uses horizontal image flipping test time augmentation to maximize the recall of the image proposal.
For the test time augmentation, we average the score heatmap and other attributes maps before decoding them into the 3D bounding box following~\cite{wang2021fcos3d}.
The invalid image proposals with lower than 0.05 class confidence are filtered before multiplying depth confidence as Eq. \ref{eq:confidence}.

Following GRIF Net~\cite{kim2020grif}, we accumulate six radar sweeps collected by multiple time frames, which is approximately 0.5 seconds.
Specifically, the location of radar points measured at time $t-\tau$ is transformed to $t$ using ego-motion, and the moving effect of surrounding objects is also compensated using the Doppler information of each radar point as:
\begin{equation}
\begin{aligned}
&x_t = x_{t-\tau} + \Delta d_{x, t, t-\tau} + v_{x, t-\tau} \cdot \Delta \tau \\
&y_t = y_{t-\tau} + \Delta d_{y, t, t-\tau} + v_{y, t-\tau} \cdot \Delta \tau \\
&z_t = z_{t-\tau} + \Delta d_{z, t, t-\tau},
\label{eq:radar accumulation}
\end{aligned}
\end{equation}
where $x_t,y_t,z_t$ denote the location of radar at time $t$, $\Delta d_{t,t-\tau}$ is the translation of ego-motion between $t$ and $\tau$, and $v$ is the ego-motion compensated radar Doppler velocity.
We take radar points within 55$m$ as input, considering the desired detection range of nuScenes is 50$m$.
If a number of accumulated radar points are more/fewer than the maximum number of radar points input $K$, we randomly sample/duplicate points.
RCS and compensated radar velocity are used as radar features after normalization.

For radial threshold (Eq. \ref{eq:radial threshold}) in Soft Polar Association (SPA), we use $\gamma=5$ for minimum association range and $\delta=10$ for modulation.
For detection heads, we use the fusion score threshold of 0.3, and the final score is computed as the geometric average of the image proposal score, fusion score, and center-ness score.
For proposals with a fusion score higher than the threshold, the predicted offset is added to the proposal center, and the image proposal velocity is replaced by the predicted velocity.
Image proposals without radar association or with a low fusion score are used as the final predictions as is.

\subsection{Loss}
Our total loss is formulated as:
\begin{equation}
\mathcal{L}_{tot}=\frac{1}{M\cdot K'}\sum_{m,i=1}^{M,K^{'}}{\mathcal{L}_{pts}} + \frac{1}{M}\sum_{m=1}^{M}{\mathcal{L}_{det}},
\label{eq:total loss}
\end{equation}
where $M$, $K^{'}$ denote the number of image proposals and associated radar points, $\mathcal{L}_{pts}$ is the binary cross-entropy loss for an auxiliary point-wise classification task.
$\mathcal{L}_{det}$ is the weighted sum of classification and regression losses described in Sec. \ref{chap:loss}.
Specifically, $\mathcal{L}_{det}$ is defined as:
\begin{equation}
\mathcal{L}_{det}=\mathcal{L}_{cls} + \mathbbm{1}_m \left[ \mathcal{L}_{cness} + \mathcal{L}_{loc} + \mathcal{L}_{vel} \right],
\label{eq:detection loss}
\end{equation}
where $\mathbbm{1}_m$ indicates whether an image proposal contains at least one valid radar return from object.
Image proposal without any radar association or none of associated radar points is reflected from the object is not used for regression loss.
$\mathcal{L}_{cls}$ and $\mathcal{L}_{cness}$ are the cross-entropy for fusion score and center-ness, $\mathcal{L}_{loc}$ and $\mathcal{L}_{vel}$ are the L1 loss for offset and velocity regression.

\section{Detailed Experimental Settings} \label{chap:appendix C}
When training the CRAFT-I, we employ an AdamW~\cite{loshchilov2018decoupled} optimizer for 140 epochs using a step learning rate scheduler decayed by 0.1 at the 90th and 120th epochs following CenterNet~\cite{Zhou2019}.
A batch size of 64, an initial learning rate of $2.5\times10^{-4}$, a weight decay of 0.0001, and gradient clipping with a max norm of 35 are used similar to~\cite{wang2021fcos3d}.

Data augmentation in multi-modality methods is non-trivial due to the coordinates discrepancy and the geometrical consistency between the two modalities.
Thanks to our image proposals and radar points fusion scheme in polar coordinate, we can apply aggressive modality-specific data augmentation strategies to the camera 3D detector, radar backbone, and fusion module.
Specifically, we apply common schemes including horizontal flipping, crop and resize, and color jittering to images when training the camera 3D detector.
During training fusion modules, we disable image scaling and cropping augmentation, which can break the 3D geometry and only apply flipping and jittering.
Instead, we first augment radar points with conventional point augmentation strategies such as rotation, translation, scaling, and flipping before feeding them into the radar backbone.
After radar feature extraction, radar points are transformed back to the original coordinate system, then associated with image proposals that are in the same coordinates.
Finally, we apply identical point augmentation strategies to both image proposals and radar points.
Additionally, we randomly change the number of radar sweeps, and adopt CBGS~\cite{zhu2019class} to handle the class-imbalance.

\section{Additional Experimental Results} \label{chap:appendix D}

\subsection{Additional State-of-the-Art Comparison}
We provide detailed results of 3D object detection on nuScenes \texttt{test} set in Table \ref{table:ns additional test set}.
We further provide our CRAFT-I and CRAFT results on nuScenes \texttt{val} set for a thorough comparison with previous state-of-the-art methods.
Some methods using DD3D depth pre-training~\cite{park2021pseudo} show 3-5\% NDS and mAP performance improvements over FCOS3D initialization~\cite{wang2021fcos3d}, where the major performance gain comes from the improved localization (mATE).
However, our CRAFT still yields the lowest translation error among DD3D pre-trained methods and even outperforms previous LiDAR method PointPillars~\cite{lang2019pointpillars}.
BEVFormer~\cite{li2022bevformer} further improves the single frame model BEVFormer-S using four consecutive images, which yields 7\% NDS and 3-4\% mAP gains which benefit from the significantly reduced velocity error.
It shows that multi-sequence image models well estimate the velocity using temporal clues and a future improvement in CRAFT is to utilize temporal information.

\setlength{\tabcolsep}{0.4em}
\setcounter{table}{6}
\begin{table*}[!t]
    \caption{
    Additional per-class analysis on nuScenes \texttt{val} set.
    We report the average number of samples in a frame (Num. GT) and the average number of radar points reflected from object (Num. Points).
    GT w/ Points denotes the percentage of samples that contain at least one radar point inside the 3D bounding box, and Image Recall is the maximum recall of CRAFT-I.
    (Valid) Associated denotes the percentage of image proposals with (valid) associated radar points.
    }
    \begin{center}
    \resizebox{0.88\textwidth}{!}{
    \begin{tabular}{c||cccccccccc}
        \hline
                   &Car&Truck&Bus&Trailer&C.V.&Ped.&M.C.&Bicycle&T.C.&Barrier\\
        \hline
        Num. GT      &11.73&2.16&0.42&0.54&0.35&5.16&0.41&0.40&2.64&4.52 \\
        \hline
        Num. Points  & 97.5&125.8&89.0&38.3&30.2&20.5&12.0&8.0&8.7&12.3 \\
        \hline
        GT w/ Points & 84.1&93.0&94.6&95.3&92.0&63.5&77.3&76.5&53.4&69.9 \\
        Image Recall & 85.6&68.6&72.8&52.0&30.7&78.2&70.5&65.9&86.9&81.2 \\
        Associated   & 80.8&67.2&70.9&51.7&30.3&69.6&65.4&64.6&70.0&72.4 \\
        Valid Assoc. & 73.7&64.9&68.4&48.8&29.2&44.6&53.8&49.0&40.2&48.3 \\
        \hline
        CRAFT-I & 52.4&25.7&30.0&15.8&5.4&39.3&28.6&29.8&57.5&47.8\\
        CRAFT  & 
        69.6\scriptsize{(\textbf{+17.2})}&37.6\scriptsize{(\textbf{+11.9})}&47.3\scriptsize{(\textbf{+17.3})}&20.1\scriptsize{(+4.3)}&10.7\scriptsize{(+5.3)}&
        46.2\scriptsize{(+6.9)}&39.5\scriptsize{(\textbf{+10.9})}&31.0\scriptsize{(+1.2)}&57.1\scriptsize{(-0.4)}&51.1\scriptsize{(+3.3)}\\
        \hline
    \end{tabular}}
    \end{center}
    \label{table:additional per class results}
\end{table*}

\subsection{Performance with Different Thresholds}
Additional results of Car class with different distance thresholds are provided in Table \ref{table:additional per distance results}.
Note that True Positive metrics (\textit{i.e.}, ATE, ASE, AOE, AVE, AAE) are calculated using a 2$m$ distance thresholds as defined in nuScenes.
We compare our method with other methods on \texttt{test} set and show the improvement over our camera-only baseline CRAFT-I on \texttt{val} set.

Although radar-only methods focus on detecting Car class, their performances are inferior to LiDAR-only and camera-only methods due to the noisy measurement of radar.
CRAFT achieves significant performance gain especially on strict matching thresholds (0.5$m$ and 1$m$) compared to previous camera-radar method and camera-only methods with the help of radar measurement.
Compared to the performance of CenterFusion is degraded from CenterNet baseline at 2$m$ and 4$m$ thresholds, our method consistently improves the CRAFT-I baseline at all thresholds.
Specifically, ours improves camera-only baseline by \textbf{32.3\%} and \textbf{23.4\%} and outperforms previous camera-radar state-of-the-art by \textbf{28.1\%} and \textbf{21.2\%} at 0.5$m$ and 1$m$ distance thresholds.
CRAFT also achieves comparable performance with PointPillars even at strict distance thresholds, which demonstrates the potential that camera and radar can replace LiDAR for autonomous driving.

\subsection{Additional Per-Class Analysis}
We analyze the statistics of annotations, radar points, and CRAFT-I on nuScenes \texttt{val} set in Table \ref{table:additional per class results}.
Although nuScenes requires shorter detection ranges for smaller objects (\textit{e.g.}, barrier and traffic cone are 30$m$, pedestrian is 40$m$, car is 50$m$), we analyze all annotations within 55$m$ for thorough analysis.
Note that we count an image proposal within 4$m$ distance to the ground truth as a true detection.

Due to the fine-grained classes in nuScenes, the distribution of classes is imbalanced.
Besides, each class has different characteristics and is often located in specific places.
For example, trailers are frequently parked near buildings and occluded by themselves (as shown in Fig. \ref{fig:additional vis} middle right), while buses are usually observed on the road.
Trailers contain much fewer radar points than buses for this reason (38.3 vs. 89.0) although they have similar sizes, lead to less significant improvement.
Similarly, construction vehicles are often occluded by barriers or walls, thus having much fewer radar points than other vehicle classes.

Radar points are often not reflected on non-metallic objects (pedestrian, bicycle, traffic cone, and barrier); hence the number of radar points return are very few.
We also assume that these classes are commonly located on the sidewalk, where more clutters exist.
Thus, these are more difficult to detect than motorcycles on the road.
In case of traffic cone, 70\% of image proposals have radar associations, but only 57\% (=40.2/70.0) of them are valid associations and others are all clutters.

\setlength{\tabcolsep}{0.4em}
\setcounter{table}{7}
\begin{table}[!t]
    \caption{
    Ablation of detection head components.}
    \begin{center}
    \resizebox{0.9\columnwidth}{!}{
    \begin{tabular}{c||c|ccccc}
        \hline
                & mAP$\uparrow$ & ATE$\downarrow$ & ASE$\downarrow$ & AOE$\downarrow$ & AVE$\downarrow$ & AAE$\downarrow$ \\
        \hline
        CRAFT-I & 52.4 & 0.49 & 0.16 & 0.07 & 1.16 & 0.15 \\
        CRAFT   & 69.6 & 0.26 & 0.16 & 0.06 & 0.31 & 0.15 \\
        \hline
        - loc   & 53.1 & 0.46 & 0.16 & 0.06 & 0.40 & 0.15 \\
        - vel   & 65.4 & 0.32 & 0.16 & 0.06 & 1.21 & 0.15 \\
        + dim   & 65.1 & 0.32 & 0.16 & 0.06 & 0.43 & 0.15 \\
        + rot   & 65.7 & 0.30 & 0.16 & 0.06 & 0.41 & 0.15 \\
        \hline
    \end{tabular}}
    \end{center}
    \label{table:additional ablation}
\end{table}

\subsection{Additional Experiments for Design Choices}
We ablate the detection head components by adding and removing the regression layers in Table \ref{table:additional ablation}.
As a default setting, we use CRAFT which has a location offset and velocity prediction heads for regression.
For dimension head, we define a 3D anchor for each class and predict size offset following~\cite{lang2019pointpillars} since the anchor-free approach fails to converge in our setting.
For rotation head, we adopt the same method in CenterNet in BEV space.

Removing the offset and velocity head significantly degrades the localization and velocity estimation performance, which shows the effectiveness of each regression head.
However, adding dimension and rotation head degrades the performance rather than improving it.
We assume that the irregular and shapeless radar points are not able to provide useful information for orientation and size estimation.
We leave a method for improving rotation and size using radar as future work.

\setcounter{figure}{5}
\begin{figure*}[t]
\begin{center}
\includegraphics[width=0.9\textwidth]{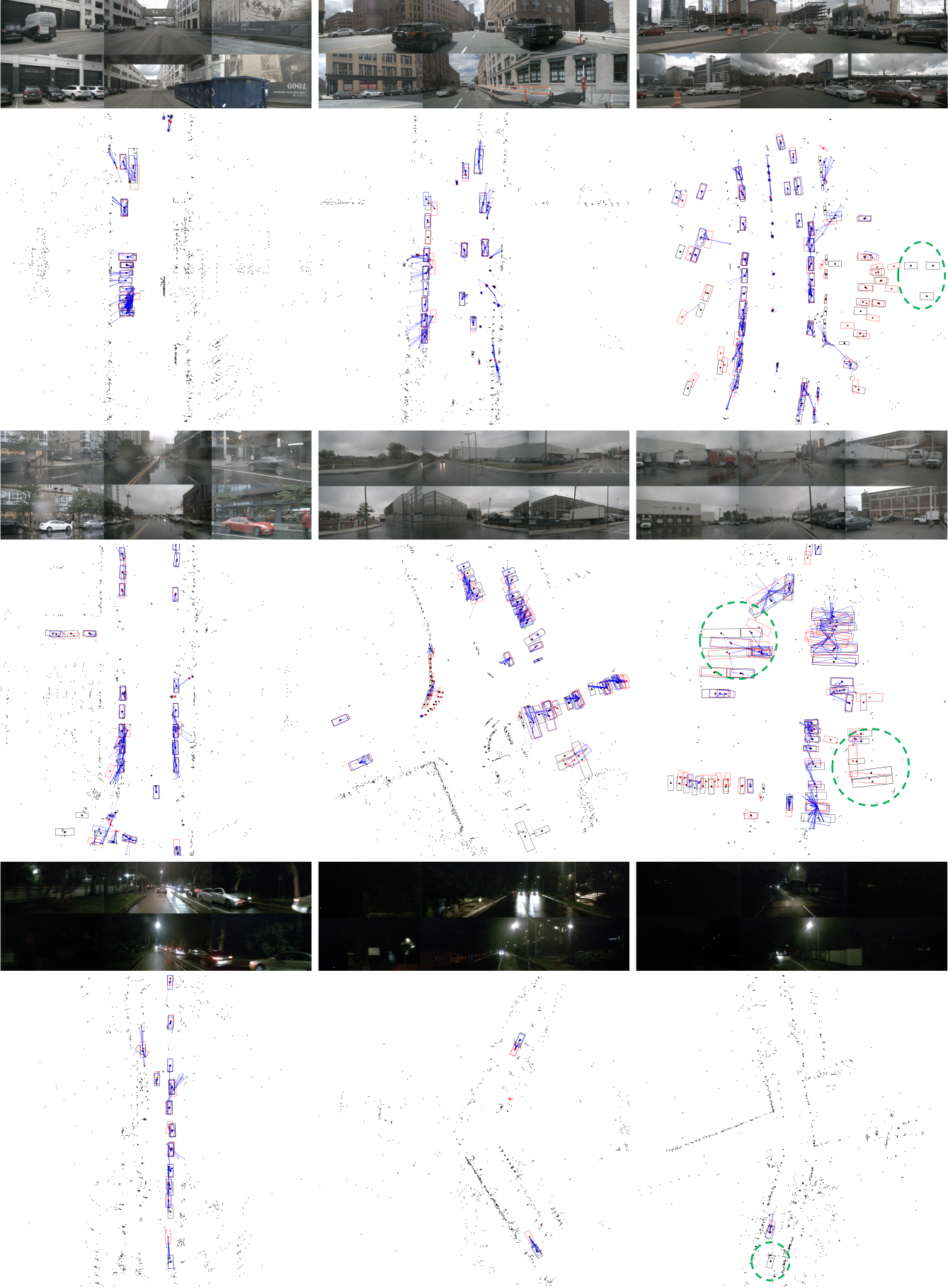}
\end{center}
\caption{
Additional qualitative results on complex (top), rainy (middle), and night (bottom) driving environments.
Black, red, and blue boxes denote the ground truth, camera-only, and fused prediction and green circles highlight some failure cases.
Note that the visualization range is set to 55m and best viewed in color with zoom in.}
\label{fig:additional vis}
\end{figure*}

\setcounter{figure}{6}
\begin{figure*}[t]
\begin{center}
\includegraphics[width=0.87\textwidth]{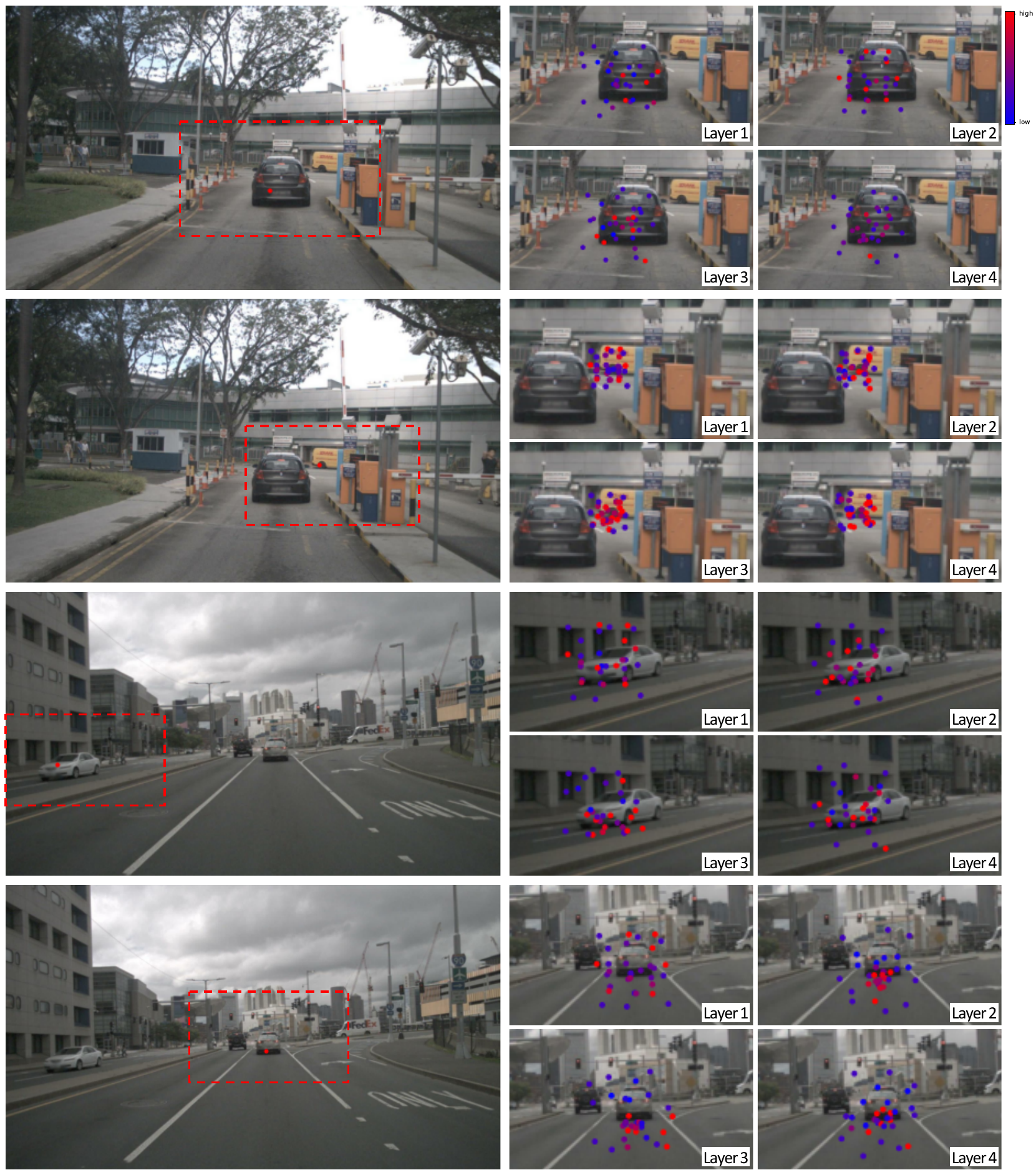}
\end{center}
\caption{
Visualization of attention in Image-to-Radar Feature Encoder.
The reference point (projected radar pixel) is shown as a red point on the left, and key sampling points (patched image features) with attention weights are shown on the right.
}
\label{fig:enc vis}
\end{figure*}

\setcounter{figure}{7}
\begin{figure*}[t]
\begin{center}
\includegraphics[width=0.72\textwidth]{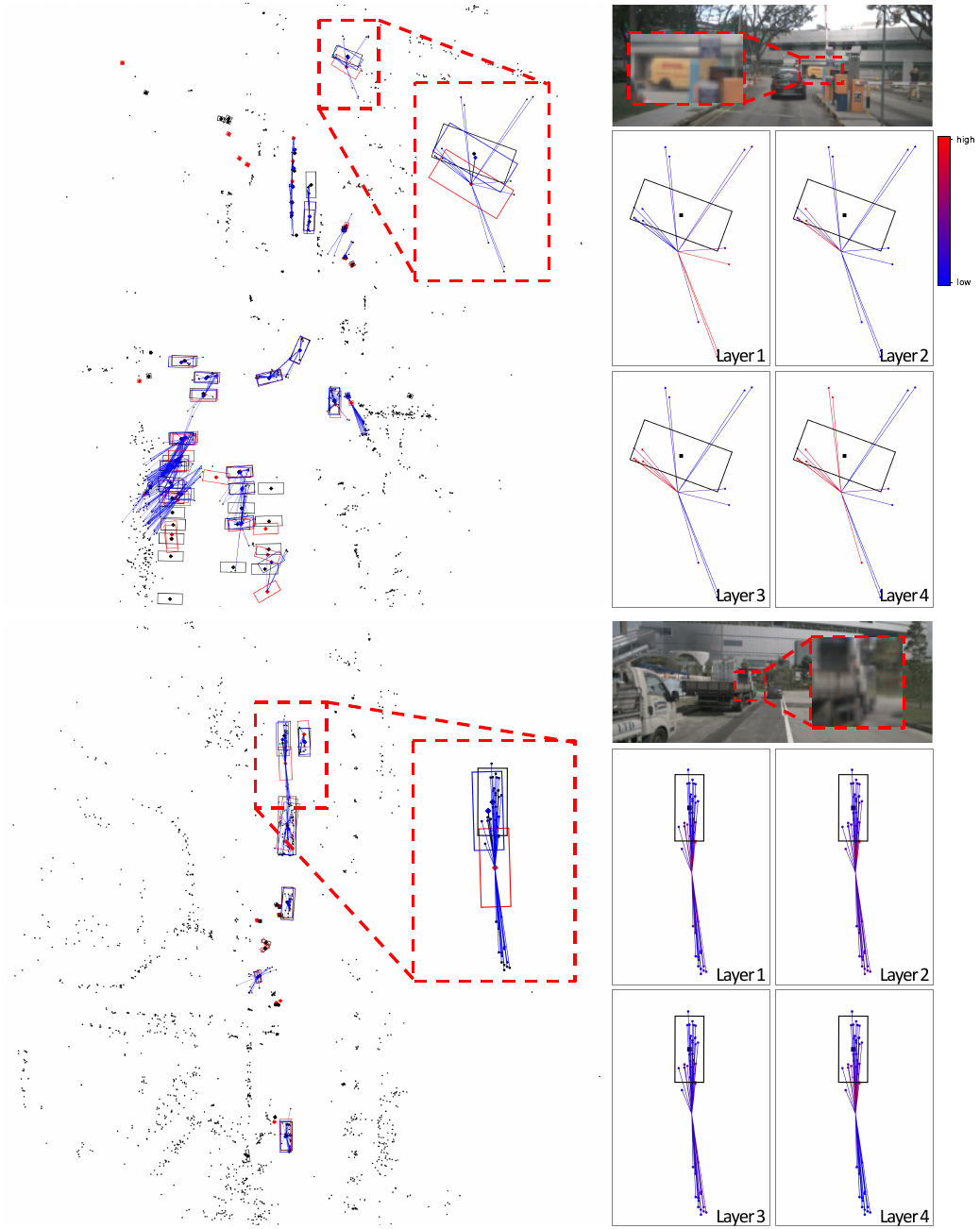}
\end{center}
\caption{
Visualization of attention in Radar-to-Image Feature Encoder.
Black, red, and blue boxes on the left denote the ground truth, camera-only, and fused prediction.
Lines between the center of the image proposal and radar points on the right indicate attention weights assigned to radar points.
}
\label{fig:dec vis}
\end{figure*}


\section{Additional Qualitative Results} \label{chap:appendix E}
We visualize additional qualitative results of CRAFT in Fig. \ref{fig:additional vis}.
These results demonstrate that our fusion method is able to robustly and accurately detect objects in various driving environments.
In most cases, camera-only predictions with inaccurate localization become accurately localized by fusing with radar.
Especially, the localization performance of using camera-only is degraded at night but significantly improved by fusion (bottom).
There are some failure modes on difficult samples that are not detected by the camera-only detector or less improved by radar fusion.
Detecting cars faraway and behind the fence (right top) or vehicles with dimming headlamps (right bottom) are more challenging and leads to false negatives.
Another common failure case is the heavily occluded trailer with only a few radar points return (right middle).

We further provide the visualization of attention in I2R and R2I feature encoder in Fig. \ref{fig:enc vis} and Fig. \ref{fig:dec vis}.
The visualization demonstrates that our spatio-contextual fusion transformer can adaptively fuse radar and image features by overcoming the noisy and ambiguous radar measurements.
In Fig. \ref{fig:enc vis}, I2R FE gradually focuses to object pixels and extracts semantic features from them.
In Fig. \ref{fig:dec vis}, given softly associated radar points, R2I FE attends more to object of interest and accurately predicts the offset to the object center.

\end{document}